\documentclass[sigconf,natbib=true]{acmart}

\usepackage{url}            
\usepackage{booktabs}       
\usepackage{amsfonts}       
\usepackage{amsmath}
\usepackage{mathtools}
\usepackage{amsthm}
\usepackage{multirow}
\usepackage{hyperref}
\usepackage[table,dvipsnames]{xcolor}
\usepackage{subcaption}
\usepackage{makecell}
\definecolor{deepgreen}{RGB}{0, 100, 0}
\usepackage{caption} 
\usepackage{pifont}

\definecolor{light-light-gray}{gray}{0.92} 
\usepackage{enumitem}
\usepackage{colortbl}
\usepackage{amsmath}
\usepackage{fvextra}
\DefineVerbatimEnvironment{VerbatimWrap}{Verbatim}{
breaklines=true, 
breakindent=0pt, 
breaksymbol={},
fontsize=\small,
}

\newcommand{\header}[1]{\vspace{2mm}\noindent\textbf{#1}}

\usepackage{newfloat}
\usepackage{listings}
\DeclareCaptionStyle{ruled}{labelfont=normalfont,labelsep=colon,strut=off} 
\floatstyle{ruled}
\newfloat{listing}{tb}{lst}{}
\floatname{listing}{Listing}
\usepackage{natbib}
\usepackage{csquotes}
\usepackage{url}
\usepackage{svg}
\usepackage{bm}
\usepackage{graphics}
\usepackage{graphicx}
\usepackage{array}
\usepackage[english]{babel}
\usepackage{textcomp}
\usepackage{latexsym}
\usepackage{siunitx}  
\usepackage{titletoc}

\usepackage{tikz}
\usetikzlibrary{tikzmark}
\usepackage[most]{tcolorbox}

\definecolor{myred}{rgb}{0.7, 0.3, 0.0}
\definecolor{myblue}{rgb}{0.2, 0.3, 0.6}
\definecolor{mygreen}{HTML}{008000}

\setcopyright{none}
\renewcommand{\footnotetextcopyrightpermission}[1]{}

\AtBeginDocument{%
  }
\usepackage{amsmath}
\usepackage{booktabs}
\usepackage{tabularx}
\usepackage{multirow}
\usepackage{makecell}
\usepackage{listings}
\usepackage{tikz}
\usepackage{xcolor}
\usepackage{pgfplots}
\pgfplotsset{compat=1.9}
\usepgfplotslibrary{patchplots}
\usepgfplotslibrary{groupplots}
\usetikzlibrary{fit, shapes.geometric, patterns, positioning, calc}

\usepackage{caption}
\usepackage{subcaption}

\usepackage{listings}

\usepackage{algorithm}
\usepackage{algpseudocode}

\newcommand{\ours}{LaSER}

\begin{document}

\title{LaSER: Internalizing Explicit Reasoning into Latent Space for Dense Retrieval}
\author{
  \textbf{Jiajie Jin}\textsuperscript{1,2} \quad
  \textbf{Yanzhao Zhang}\textsuperscript{2} \quad
  \textbf{Mingxin Li}\textsuperscript{2} \quad
  \textbf{Dingkun Long}\textsuperscript{2}\\
  \textbf{Pengjun Xie}\textsuperscript{2} \quad
  \textbf{Yutao Zhu}\textsuperscript{1} \quad
  \textbf{Zhicheng Dou}\textsuperscript{1}
}
\affiliation{
  \institution{\textsuperscript{1}Gaoling School of Artificial Intelligence, Renmin University of China}
  \country{}
  \institution{\textsuperscript{2}Tongyi Lab, Alibaba Group}
  \country{}
}
\authornote{Corresponding author.}







\renewcommand{\shortauthors}{Jiajie Jin et al.}

\begin{abstract} 
Large Language Models (LLMs) have fundamentally transformed dense retrieval, upgrading backbones from discriminative encoders to generative architectures. However, a critical disconnect remains: while LLMs possess strong reasoning capabilities, current retrievers predominantly utilize them as static encoders, leaving their potential for complex reasoning unexplored. To address this, existing approaches typically adopt ``rewrite-then-retrieve'' pipelines to generate explicit Chain-of-Thought (CoT) rationales before retrieval. However, this incurs prohibitive latency. Conversely, implicit reasoning methods utilizing latent tokens offer efficiency but often suffer from semantic degeneration due to the lack of explicit supervision. In this paper, we propose \textbf{\ours{}}, a novel self-distillation framework that internalizes explicit reasoning into the latent space of dense retrievers. Operating on a shared LLM backbone, \ours{} introduces a dual-view training mechanism: an Explicit view that explicitly encodes ground-truth reasoning paths, and a Latent view that performs implicit latent thinking. To bridge the gap between these views, we design a multi-grained alignment strategy. Beyond standard output alignment, we introduce a \textit{trajectory alignment} mechanism that synchronizes the intermediate latent states of the latent path with the semantic progression of the explicit reasoning segments. This allows the retriever to ``think'' silently and effectively without autoregressive text generation. Extensive experiments on both in-domain and out-of-domain reasoning-intensive benchmarks demonstrate that \ours{} significantly outperforms state-of-the-art baselines. Furthermore, analyses across diverse backbones and model scales validate the robustness of our approach, confirming that our unified learning framework is essential for eliciting effective latent thinking. Our method successfully combines the reasoning depth of explicit CoT pipelines with the inference efficiency of standard dense retrievers\footnote{The code, model, and training data are available at \url{https://github.com/ignorejjj/LaSER}.}. 

\end{abstract}



\maketitle

\section{Introduction}

Dense retrieval has evolved into a fundamental component of modern Information Retrieval (IR) systems, widely deployed in web search, question answering~\cite{hotpotqa, dpr}, and Retrieval-Augmented Generation (RAG) systems~\cite{rag, ragsurvey}. These retrievers effectively bridge the vocabulary mismatch gap via semantic matching in a shared embedding space~\cite{dpr,bge,e5}. With the advent of Large Language Models (LLMs)~\cite{llm-survey,llm4ir,ir_meets_llm}, the field has witnessed a critical paradigm shift: retriever backbones are transitioning from small-scale discriminative models (e.g., BERT, RoBERTa) to large-scale generative LLMs~\cite{e5mistral,gte}. State-of-the-art LLM-based retrievers~\cite{nvembed, qwen3embedding,bge} have demonstrated superior semantic understanding capabilities compared to their predecessors. 
While LLM-based retrievers have demonstrated superior semantic understanding capabilities~\cite{e5mistral, qwen3embedding,bge_m3}, a paradox remains: while these retrievers inherit the massive knowledge and reasoning potential of the underlying LLMs, they are primarily trained via contrastive learning to optimize representational distinctiveness~\cite{infonce}. This training objective largely leaves the inherent reasoning capabilities of the generative architecture dormant, treating the LLM merely as a stronger encoder rather than a reasoner~\cite{brightbench,reasonir,gircse}.

\begin{figure*}[!t]
    \centering
    \includegraphics[width=\linewidth]{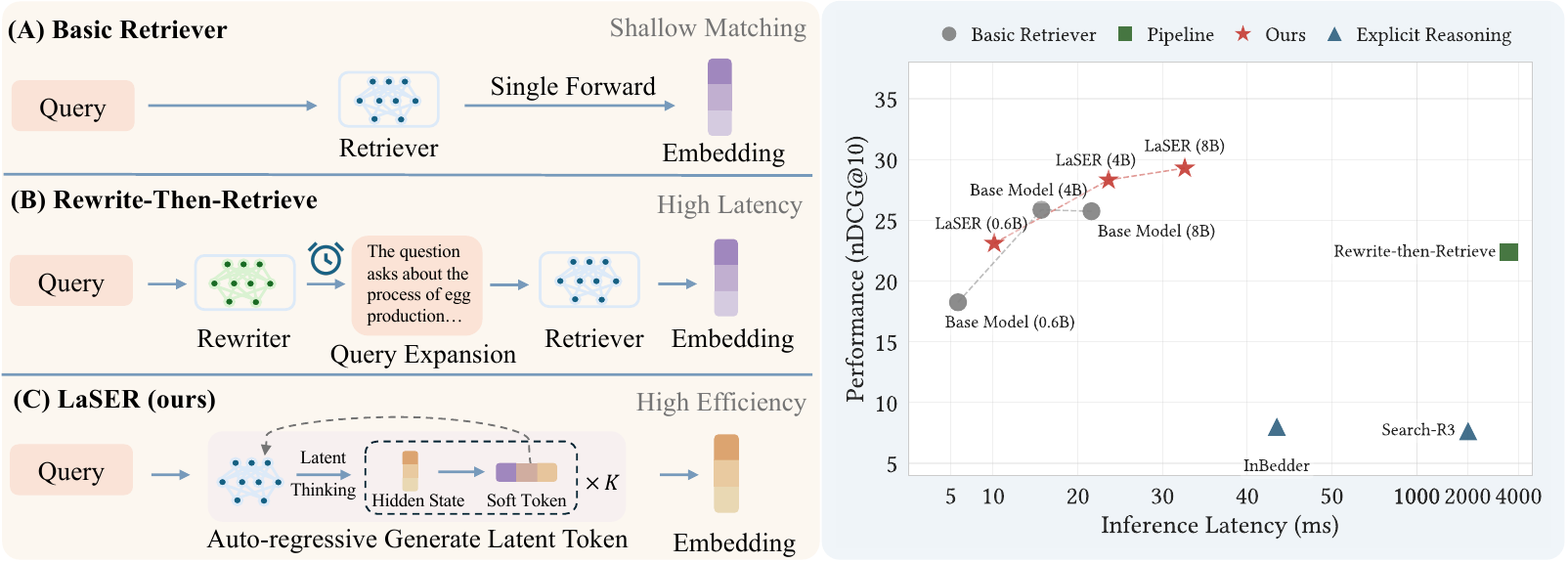}
    \caption{
    \textbf{Left:} Comparison between our proposed pipeline and conventional retriever and rewrite-then-retrieve method. \textbf{Right:} Comparison of performance and latency of different methods on the Bright benchmark. Letency represents the time consumption of processing a single query. The ``Rewrite-then-Retrieve'' method uses a 0.6B retriever and a 3B rewriter.
    }
    \label{fig:intro_fig}
\end{figure*}

This limitation becomes particularly evident when handling complex user queries. Real-world search scenarios often involve implicit intents~\cite{intent_demonstrations_zhao,conv_search_survey}, multi-hop logic~\cite{hotpotqa,2wiki}, or ambiguous descriptions~\cite{ambigqa, followir} that require reasoning beyond superficial semantic matching. To bridge this gap, a prevalent solution adopts a ``rewrite-then-retrieve'' pipeline~\cite{hyde, conv_search_rewrite}, utilizing external LLMs to explicitly generate query expansions or Chain-of-Thought (CoT) rationales before retrieval~\cite{COT,think_then_embed,expandr, grace,zhang_secretlyreasoner}. 
While effective in enhancing performance, their dependency on verbose autoregressive generation creates prohibitive latency bottlenecks~\cite{search-r3} (as shown in Figure~\ref{fig:intro_fig}).
To mitigate these latency issues, recent studies have explored \textit{implicit reasoning} approaches, allowing retrievers to generate ``latent thinking tokens'' within the embedding space before producing the final representation~\cite{gircse}. While promising, these methods face a critical challenge: \textbf{the lack of explicit semantic supervision.} Relying solely on the final contrastive learning loss, the generated latent tokens often lack sufficient guidance to accurately capture the critical information required for retrieval, particularly for queries involving reasoning.

We argue that \textbf{the key to resolving this dilemma lies in internalization}: leveraging the explicit reasoning capabilities of LLMs as privileged supervision to guide the lightweight latent thinking of the retriever. Our core insight is that while generating explicit CoT during inference is inefficient, the semantics carried by these reasoning paths are invaluable. Instead of treating explicit and implicit reasoning as separate paradigms, we propose to align them within a unified framework, allowing the retriever to ``think'' silently in latent space.
Driven by this insight, we propose a dual-view framework where the model learns to internalizes explicit reasoning capabilities within latent space. The Explicit-View learns to encoding the reasoning logic into embedding, serving as a semantic upper bound. The \textbf{Latent-View} generates latent thinking tokens to simulate the reasoning process internally.

To effectively transfer reasoning capabilities from the explicit view to the latent view, we introduce a multi-grained alignment strategy. We align not only the final query representations (output alignment) but also apply auxiliary supervision to intermediate latent states (trajectory alignment), encouraging the latent tokens to capture the semantic progression of the explicit CoT.
This framework offers significant versatility. During training, the explicit view acts as a semantic teacher; during inference, \ours{} operates in a streamlined mode using latent reasoning, eliminating the need for text generation while maintaining performance comparable to computationally intensive pipeline methods. 
We evaluate \ours{} on a comprehensive suite of embedding benchmarks, focusing on reasoning-intensive tasks such as Bright~\cite{brightbench}, BrowseComp-Plus~\cite{BrowseComp-plus}, and FollowIR~\cite{followir}. Empirical results demonstrate that our method significantly enhances retrieval performance by introducing only a few latent tokens during inference, avoiding the latency of autoregressive decoding. Furthermore, \ours{} consistently outperforms existing explicit and implicit reasoning baselines. Notably, in certain reasoning-intensive scenarios, it even achieves superior performance compared to computation-heavy pipelines that rely on external LLMs for query rewriting, validating the effectiveness of our proposed distillation mechanism. 

Our main contributions are summarized as follows:
\begin{itemize}[leftmargin=1em]
    \item We propose \ours{}, a unified retrieval framework that bridges the gap between explicit CoT reasoning and implicit latent thinking via self-distillation within a shared backbone.
    \item We design a dual-view alignment mechanism, featuring both output-level and process-level trajectory alignment, which effectively distills reasoning semantics into latent tokens and prevents the degeneration often observed in implicit reasoning models.
    \item Extensive experiments demonstrate that \ours{} consistently outperforms baselines across diverse backbone architectures and model scales, robustly validating the efficacy of our proposed self-distillation mechanism. 
    \item We empirically validate that the semantics of explicit reasoning paths can be effectively distilled into the latent space. This allows \ours{} to achieve performance comparable to computationally intensive "rewrite-then-retrieve" pipelines while maintaining the inference efficiency of standard dense retrievers.
\end{itemize}

\section{Related Work}

\subsection{LLM-Based Retrieval and Query Expansion} 
The paradigm of dense retrieval has shifted significantly from small-scale discriminative backbones (e.g., BERT~\cite{bert}) to large-scale generative LLMs~\cite{e5mistral, qwen3embedding}. By leveraging the massive world knowledge and semantic understanding inherent in LLMs, these retrievers achieve superior performance in standard semantic matching. However, they continue to rely primarily on contrastive learning objectives, which often struggle with complex queries requiring deep reasoning or intent disambiguation. To bridge this gap, the ``rewrite-then-retrieve'' pipeline has become prevalent (as shown in Figure~\ref{fig:intro_fig}), utilizing external LLMs to expand queries~\cite{hyde, query2doc} or generate reasoning rationales prior to retrieval~\cite{reasonembed, tongsearch}. While effective, this approach decouples reasoning from retrieval, introducing significant latency due to autoregressive decoding and complicating system deployment. Unlike these multi-stage pipelines, our work aims to \textit{internalize} the rewriting and reasoning capabilities directly into the retriever’s parameters, preserving the inference efficiency of a single-stage dual-encoder while retaining the semantic benefits of generative expansion.

\subsection{Reasoning in Retrieval} Recent research has explored integrating reasoning capabilities directly into retrieval systems, generally categorized into data-centric~\cite{reasonrank, rader,bge} and mechanism-centric approaches. Data-centric methods enhance retrieval by synthesizing reasoning-intensive queries with corresponding hard negatives. Mechanism-centric approaches, conversely, modify the architecture of retriever to support reasoning, further dividing into \textit{explicit} and \textit{implicit} strategies. Explicit methods leverage the retriever itself to generate natural language CoT rationales before producing the final embedding~\cite{grace,expandr,think_then_embed,search-r3,inbedder}. While this approach improves interpretability, it complicates training by requiring the model to learn two tasks simultaneously~\cite{2402_gritlm}. Furthermore, it suffers from the same computational bottlenecks as external rewriting. Implicit reasoning, alternatively, simulates this process within the embedding space by generating ``latent thinking tokens''~\cite{gircse} or multi-vector representations~\cite{diverse_multiquery_retrieval} that capture query nuances without visible text generation. Despite their efficiency, existing implicit methods predominantly rely on the final contrastive loss for supervision, which acts as a bottleneck for capturing complex query semantics and hinders further performance improvements.
Our framework addresses this limitation by using explicit CoT as a privileged supervisor, distilling the semantics of explicit reasoning paths into latent tokens to ensure the implicit process is semantically meaningful.

\subsection{Knowledge Distillation in IR}
Knowledge Distillation (KD) is widely used to transfer capabilities from powerful teachers to efficient students~\cite{knowledge_distillation_survey}, generally fall into \textit{output-level} and \textit{feature-level} distillation in IR. Output-level distillation typically utilizes LLMs to generate synthetic training data, such as query expansions or pseudo-relevance labels~\cite{qwen3embedding}, which the student mimics. Feature-level distillation focuses on aligning continuous representations, commonly training a student retriever to approximate the hidden space or relevance scores of a powerful teacher model~\cite{distillation_sota,adaptive_distillation}.
Critically, these methods are outcome-oriented: they improve relevance signals but do not fundamentally alter the student's information processing mechanism. Distinct from this paradigm, we draw inspiration from LLM latent reasoning~\cite{latent_cot_survey, coconut, codi} to internalize explicit reasoning steps into latent states~\cite{gircse}. Instead of merely distilling relevance scores, we leverage explicit Chain-of-Thought as ``ground truth'' logic to supervise the latent thinking. To our knowledge, \ours{} the first work to apply explicit-to-latent reasoning distillation in dense retrieval.

\section{Methodology}

\begin{figure*}[!t]
    \centering
    \includegraphics[width=\linewidth]{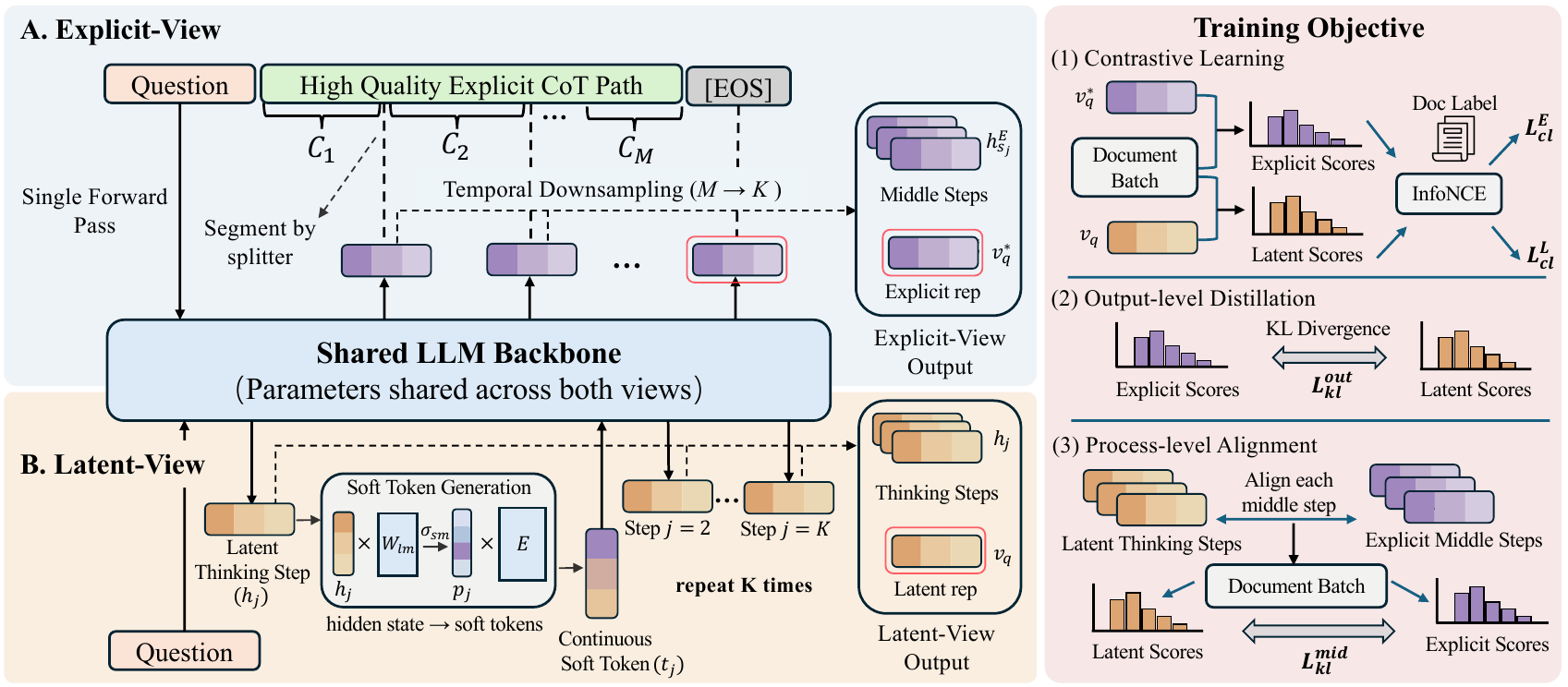}
    \caption{
    Illustration of the \ours{}. During training, each step involves two inference pathways: the explicit-view and the latent-view (shown on the left). The inputs for these two paths differ, with the explicit-view incorporating additional information. However, during the inference phase, only the latent-view mode is utilized. The loss function is computed based on the results from both paths (shown on the right).
    }
    \label{fig:main_arch}
\end{figure*}

In this section, we present \ours{}, a unified framework designed to internalize explicit reasoning capabilities into the latent space of dense retrievers. We first formulate the retrieval problem and give an overview of our framework. Then, we detail the training process and optimization objectives of our framework, highlighting our multi-grained self-distillation mechanism.

\subsection{Problem Formulation}
Given a user query $q$ and a candidate document corpus $\mathcal{D} = \{d_1, d_2, \dots, d_N\}$, the goal of dense retrieval is to learn an encoder $f(\cdot)$ that maps textual inputs into a $m$-dimensional embedding space $\mathbb{R}^m$. The relevance score between a query $q$ and a document $d$ is typically computed via the cosine similarity of their embeddings $\mathit{v}_q$ and $\mathit{v}_d)$, i.e., $s(q, d) = \cos(\mathit{v}_q, \mathit{v}_d)$.

In standard LLM-based retrievers, the query representation is obtained through a single forward pass. To enhance instruction following, a task-specific instruction $I$ is often prepended to $q$. An [EOS] token is then appended to the input sequence, and the hidden state of the last layer corresponding to this token is extracted as the embedding:
\begin{equation}
    \mathit{v}_q = f([I_\text{task}; q; \text{[EOS]}]).
\end{equation}
However, this shallow processing often fails to capture the complex implicit intents required for reasoning-intensive queries. 

To address this, existing ``rewrite-then-retrieve'' approaches introduce an external LLM reasoner $\mathcal{M}$ to generate an explicit reasoning path and the retriever then encodes this enriched context:
\begin{equation}
    \mathit{v}_q^{*} = f([I_\text{task}; q; r_q; \text{[EOS]}]), \\
    \text{where}\; r_q = \mathcal{M}(I_{q2r}, q).\label{eq:vqstar}
\end{equation}
While $\mathit{v}_q^{*}$ captures deeper semantics, the autoregressive generation of textual $r_q$ incurs prohibitive latency and deployment costs.

In this work, we aim to bridge this gap by training a model to perform \textit{latent reasoning}. Instead of generating explicit text $r_q$, the model autonomously generates a sequence of $K$ continuous latent thinking tokens $T = \{\mathit{t}_1, \dots, \mathit{t}_K\}$ within the embedding space. The final query representation $\mathit{v}_q$ is derived from these latent states, aiming to approximate the semantic depth of $\mathit{v}_q^{*}$ while maintaining the efficiency of a single-model architecture.

\subsection{Overview of Framework}
As illustrated in Figure~\ref{fig:main_arch}, \ours{} is built upon a unified generative LLM backbone with causal attention. To achieve the internalization of reasoning capabilities, we employ a dual-view training paradigm where the model learns through two aligned perspectives sharing the same parameters $\theta$:

\header{Explicit-View.} This view acts as the semantic upper bound during training. It receives the query augmented with a high-quality, explicit CoT rationale generated by a superior reasoner. By performing a single forward pass on this enriched context, the explicit-view acquires a representation that comprehensively synthesizes the original query and the explicit reasoning path.

\header{Latent-View.} This view serves as the target mode for inference. It takes only the original query as input, and  generate a sequence of \textit{latent thinking tokens} in the latent embedding space before producing the final query representation.
We design a self-distillation framework to facilitate the co-training of these two views, especially the latent-view, as detailed in Section~\ref{sec:optimization}.

\header{Inference.} During inference, \ours{} operates exclusively in the \textbf{\textit{latent-view}}. It utilizes the learned latent thinking tokens to encode queries, ensuring the latency matches standard dense retrievers while benefiting from internalized reasoning. Optionally, the model retains the flexibility to accept rewritten queries from external LLMs if available, further adapting to diverse deployment scenarios.

\subsection{Latent View}
\label{sec:latent_view}

The primary objective of the Latent View is to enhance the model's reasoning capacity for complex queries while circumventing the latency overhead associated with explicit text generation. Inspired by recent advances in continuous reasoning, we propose a mechanism where the encoder model, parameterized by $\theta$, performs ``thinking'' within a continuous embedding space. This approach replaces discrete token generation with an autoregressive sequence of continuous thought vectors, allowing the model to refine its understanding of the query intent in a fully differentiable manner.

As shown in Figure~\ref{fig:main_arch}, given an input query $q$, the model first maps the discrete token sequence into dense embeddings $\mathbf{X}_0 = \mathbf{E}(q)$, where $\mathbf{E} \in \mathbb{R}^{|\mathcal{V}| \times m}$ is the embedding matrix and $|\mathcal{V}|$ is the vocabulary size. The backbone LLM $f_\theta$ processes this sequence to produce initial hidden states.

To facilitate latent reasoning, we extend the query with a sequence of $K$ latent thinking tokens $T = \{\mathit{t}_1, \dots, \mathit{t}_K\}$. Unlike standard decoding which samples discrete indices, we construct these tokens as continuous vectors to preserve semantic richness and differentiability. The process proceeds autoregressively: at the $j$-th thinking step ($1 \le j \le K$), let $\mathit{h}_{j-1} \in \mathbb{R}^m$ be the last hidden state output by the backbone. We project $\mathit{h}_{j-1}$ to the vocabulary space via the language modeling head $\mathit{W}_{\text{lm}} \in \mathbb{R}^{m \times |\mathcal{V}|}$ to obtain logits $\mathit{l}_j$:
\begin{equation}
    \mathit{l}_j = \mathit{W}_{\text{lm}} \mathit{h}_{j-1}.
\end{equation}
Rather than performing a hard selection (e.g., $\arg\max$), we compute a probability distribution $\mathit{p}_j = \text{Softmax}(\mathit{l}_j)$ over the vocabulary. The soft latent token $\mathit{t}_j$ is then computed as the expected embedding vector under this distribution:
\begin{equation}
    \mathit{t}_j = \mathit{p}_j^\top \mathbf{E} = \left(\text{Softmax}(\mathit{W}_{\text{lm}} \mathit{h}_{j-1})\right)^\top \mathbf{E}.
    \label{eq:soft_token}
\end{equation}

This soft token $\mathit{t}_j$ is appended to the input sequence for the next step. The model operates autoregressively, where the input at step $j$ includes the original query and all preceding latent tokens:
\begin{equation}
    \mathbf{H}_j = f_\theta\left([\mathbf{E}(q), \mathit{t}_1, \dots, \mathit{t}_j]\right),
\end{equation}
where $\mathbf{H}_j$ represents the sequence of hidden states. Through the causal attention mechanism, each latent token $\mathit{t}_j$ can attend to the query context and previous thinking steps, mimicking the sequential logic of explicit Chain-of-Thought reasoning.

After $K$ steps, we obtain a sequence of reasoning states $\mathcal{H}_{\text{think}} = \{\mathit{h}_1, \dots, \mathit{h}_K$\}. To synthesize the global semantics of this reasoning process into a unified representation, we apply mean-pooling over these states to derive the final query vector $\mathit{v}_q$:
\begin{equation}
    \mathit{v}_q = \frac{1}{K} \sum_{j=1}^{K} \mathit{h}_j.
\end{equation}
The representation of documents $\mathit{v}_d$ are also obtained in the same way. This latent thinking process maintains strict autoregressiveness, ensuring full compatibility with KV caching optimizations. Furthermore, by limiting the reasoning horizon $K$ (typically $K < 10$), the method achieves inference latency comparable to standard dense retrievers, significantly outperforming generative approaches that require decoding long textual chains. 

\subsection{Explicit-View}
\label{sec:explicit_view}

The Explicit-View serves as the semantic teacher, learning explicit reasoning processes through privileged knowledge derived from an advanced external reasoner. 
Specifically, for each query $q$, we prompt the external reasoner to generate a comprehensive CoT rationale $r_q$, which includes the intermediate reasoning logic and understanding of the potential intent behind the query (refer to ~\ref{tab:annot_prompt} for the detailed prompt). 
We construct the augmented input sequence $x_{\text{aug}}$ by concatenating the original query $q$, the generated rationale $r_q$, and [EOS] token:
\begin{equation}
    x_{\text{exp}} = [q; r_q; \text{[EOS]}].
\end{equation}
The model then encodes this augmented input as in Equation~\ref{eq:vqstar}.
As the inputs of the explicit-view and latent-view are different, there is no mutual interference between the two tasks. Since the reasoning logic is explicitly included in the textual input, the encoder only performs a single forward pass and get the last hidden state as the final representation $\mathit{v}_q^*$, as shown in Equation~(\ref{eq:vqstar}). Given that the explicit path itself can provide intermediate signals, we additionally extract the hidden states of certain intermediate tokens in the sequence to represent the intermediate reasoning process (as shown in Figure~\ref{fig:main_arch}). These details will be discussed in Section~\ref{sec:process_align}.

\begin{table}[t]
\centering
\small 
\caption{The prompt used for annotating the explicit reasoning path of the given query.}
\label{tab:annot_prompt}
\begin{tabularx}{\linewidth}{@{}X@{}} 
\toprule
\textbf{Instruction:} \\
Given a question, your mission is to follow the instructions below: \\
1. Identify the essential problem. \\
2. Think step by step to reason and describe what information could be relevant and helpful to address the questions in detail. \\
3. Draft an answer with as many thoughts as you have. \\
\textbf{The given question:} \\
\texttt{[Begin of Question]} \\
\{Original Query\} \\
\texttt{[End of Question]} \\
\bottomrule
\end{tabularx}
\end{table}

\subsection{Optimization via Self-Distillation}
\label{sec:optimization}
To transfer the reasoning capability from the Explicit-View to the Latent-View, we propose a joint training objective combining contrastive learning and multi-grained distillation.

\subsubsection{Contrastive Training}
Both two views must learn basic discriminative signals. We employ the standard InfoNCE loss~\cite{infonce} for both views using ground-truth positive documents $d^+$ and negatives $d^-$.
For the latent-view:
\begin{equation}
    \mathcal{L}_{cl}^{L} = -\frac{1}{N}\sum_{i=1}^N\log \frac{\exp(\mathit{v}_q \cdot \mathit{v}_{d^+} / \tau)}{\sum_{d \in \{d^+\} \cup \{d^-\}} \exp(\mathit{v}_q \cdot \mathit{v}_d / \tau)},
\end{equation}
where $N$ is the batch size, $\tau$ is the temperature hyperparameter.
A similar loss $\mathcal{L}_{cl}^{E}$ is computed for the explicit view using $\mathit{v}_q^*$.

\subsubsection{Output-Level Distillation}
\label{sec:output_distill}
To ensure the Latent-View effectively internalizes the explicit reasoning capabilities, we distill the semantic knowledge from the Explicit-View to the Latent-View. Instead of strictly aligning the query embeddings $\mathit{v}_q$ and $\mathit{v}_q^*$ , which may over-constrain the latent view given the inherent information gap between the raw and reasoning-augmented inputs, we adopt a score-based distillation strategy. This approach focuses on transferring the ranking preferences encapsulated in the explicit view. 
Formally, for a query $q$ and a document batch $\mathcal{B} = \{d_1, \dots, d_N\}$, we align the probability distributions of relevance scores derived from both views.

Let $s^E_{i} = \mathit{v}_q^* \cdot \mathit{v}_{d_i}$ and $s^L_{i} = \mathit{v}_q \cdot \mathit{v}_{d_i}$ denote the similarity scores from the Teacher and Student, respectively. We apply a softmax function with a distillation temperature $\tau_{kd}$ to obtain the probability distributions:
\begin{equation}p^E_i = \frac{\exp(s^E_{i} / \tau_{kd})}{\sum_{k=1}^N \exp(s^E_{k} / \tau_{kd})}, \quad p^L_i = \frac{\exp(s^L_{i} / \tau_{kd})}{\sum_{k=1}^N \exp(s^L_{k} / \tau_{kd})}.
\end{equation}
The output-level distillation loss is then defined as the Kullback-Leibler (KL) divergence between these two distributions:
\begin{equation}\mathcal{L}_{kl}^{out} = \frac{1}{N} \sum_{i=1}^N p^E_i \log \left( \frac{p^E_i}{p^L_i} \right).
\end{equation}
By minimizing $\mathcal{L}_{kl}^{out}$, the latent view learns to mimic the fine-grained information of document relevance in explicit view, effectively capturing the reasoning-enhanced judgment logic without requiring identical embedding coordinates. 

\subsubsection{Process-Level Trajectory Alignment}
\label{sec:process_align}
Relying solely on output alignment often treats the reasoning process as a black box, which may lead to the degeneration of latent tokens where they fail to encode meaningful intermediate semantics. To prevent this, we introduce a \textit{Trajectory Alignment} mechanism to explicitly supervises the intermediate thinking steps.

Our design operates on the hypothesis that the sequence of latent tokens should function as a compressed semantic trajectory of the verbose explicit CoT. While the explicit rationale $r$ consists of variable-length logical segments $\{s_1, ..., s_M\}$ (where $M$ is the number of reasoning steps), the latnet view operates with a fixed budget of $K$ tokens. Since typically $M \geq K$, we posit that each latent token $t_i$ should correspond to a specific semantic keyframe within the explicit reasoning path.
To realize this, we employ a \textbf{Temporal Downsampling} strategy to align the granularities of the two views.  We map the $i$-th latent step to a corresponding explicit segment index $j_i$ via uniform sampling:
\begin{equation}
    j_i = \lfloor i \times \frac{M}{K} \rfloor.
\end{equation}
This mapping ensures that the model synchronizes its latent thoughts with the semantic progression of the explicit reasoning chain. We then align the intermediate states by minimizing the KL divergence between their projected distributions over the document batch:
\begin{equation}
    \mathcal{L}_{kl}^{mid} = \frac{1}{K} \sum_{i=1}^{K} \text{KL} \left( \sigma(\mathit{h}^E_{j_i} \cdot \mathit{v}_{\mathcal{D}}), \sigma(\mathit{h}_i \cdot \mathit{v}_{\mathcal{D}}) \right),
\end{equation}
where $\sigma$ denotes the softmax function over the document batch. By enforcing this alignment, we ensure that the latent tokens don't drift but instead act as semantic checkpoints that rigorously approximate the explicit reasoning logic.

\subsubsection{Overall Objective}

The final training objective is a weighted sum of the components:
\begin{equation}
    \mathcal{L} = \mathcal{L}_{cl}^{L} + \lambda_1 \mathcal{L}_{cl}^{E} + \lambda_2 \mathcal{L}_{kl}^{out} + \lambda_3 \mathcal{L}_{kl}^{mid},
\end{equation}
where $\lambda_1,\lambda_2, \lambda_3$ are hyperparameters that used to balance the tasks.

\section{Experiments Design}

\subsection{Datasets and Metrics} 
To improve the model's reasoning capabilities, we utilize the synthetic dataset from ReasonEmb~\cite{reasonembed} for training, which includes 81k training examples across 12 domains. Each training query accompany with a reasoning path, which is generated by GPT-4o-mini~\cite{gpt_4o_system_card} as a reasoner, serving as the explicit input in our explicit-view. 

For evaluation, we select three benchmarks requiring deep reasoning: Bright~\cite{brightbench},  FollowIR~\cite{followir}(out-of-domain) and BrowseComp-Plus~\cite{BrowseComp-plus}(out-of-domain).  Following standard practices, we report nDCG@10 for BRIGHT, and Recall@5, Recall@100 and Recall@1000 for BC-Plus. For FollowIR, the evaluation for each subset includes a standard metric alongside p-MRR, a pairwise metric designed to assess instruction-following capability. Adhering to official protocols, in addition to p-MRR, we report MAP@5 for the Robust'04 and Core'17 subsets, and nDCG@5 for the News'21 subset. 

The statistics of both training datasets and evaluation datasets are shown in Table~\ref{tab:data_stat}.

\subsection{Baselines}
We compare \ours{} against four categories of baseline methods. To ensure a rigorous comparison, unless otherwise noted, all trainable baselines utilize the same backbone and training data as our method.

\begin{table}[h]
    \centering
    \setlength{\tabcolsep}{3.5pt}    
    \caption{Statistics of the datasets used in our experiments. ``Q'' and ``C'' denote Query and Corpus respectively. ``Len.'' denotes average token count. ``Reas. Len.'' denotes average reasoning length annotated by external LLM.}
    \begin{tabular}{llrrrc}
        \toprule
        \textbf{Split} & \textbf{Dataset} & \textbf{\# Q} & \textbf{\# C} & \textbf{Q Len.} & \textbf{Reas. Len.} \\
        \midrule
        Train & ReasonEmb & 81,659 & -- & 222.1 & 984.8 \\
        \midrule
        \multirow{3}{*}{Test} & Bright (ID) & 1,384 & 1,145,164 & 240.8 & 2,412.8 \\ 
        & FollowIR (OOD) & 104 & 98,312 & 76.6 & -- \\
        & BC-Plus (OOD) & 830 & 100,195 & 123.2 & -- \\
        \bottomrule
    \end{tabular}
    \label{tab:data_stat}
\end{table}

\begin{table*}[t]
\centering
\caption{Main retrieval performance on the Bright benchmark. We report nDCG@10 for all subsets. The best results in the same backbone setting are formatted in \textbf{bold} and the second best are \underline{underlined}. $\dagger$ indicates the pipeline method that uses an external LLM to rewrite queries during inference.}
\label{tab:main_results}
\resizebox{\textwidth}{!}{
\begin{tabular}{l|c|ccccccc|cc|ccc|c}
\toprule
\multirow{2}{*}{\textbf{Models}} & \multirow{2}{*}{\textbf{Size}} & \multicolumn{7}{c|}{\textbf{StackExchange}} & \multicolumn{2}{c|}{\textbf{Coding}} & \multicolumn{3}{c|}{\textbf{Theorem-based}} & \multirow{2}{*}{\textbf{Avg.}} \\
\cmidrule(lr){3-9} \cmidrule(lr){10-11} \cmidrule(lr){12-14}
& & \textbf{Bio.} & \textbf{Earth.} & \textbf{Econ.} & \textbf{Psy.} & \textbf{Rob.} & \textbf{Stack.} & \textbf{Sus.} & \textbf{Leet.} & \textbf{Pony} & \textbf{AoPS} & \textbf{TheoQ.} & \textbf{TheoT.} & \\
\midrule
\multicolumn{15}{l}{\textit{\textbf{Standard Dense Retrievers}}} \\
BGE-M3 & 0.6B & 7.2 & 13.3 & 12.5 & 12.8 & 12.6 & 10.0 & 10.1 & 15.6 & 30.7 & 1.5 & 5.6 & 5.0 & 11.4 \\
Multilingual-E5-Large-Instruct & 0.6B & 15.0 & 24.6 & 14.0 & 14.6 & 16.1 & 10.3 & 13.8 & 15.1 & 2.1 & 3.0 & 11.6 & 6.9 & 12.3 \\
E5-Mistral-7B-Instruct & 7B & 17.9 & 28.6 & 18.5 & 16.5 & 18.6 & 13.4 & 20.0 & 19.7 & 6.3 & 2.0 & 14.2 & 22.1 & 16.5 \\
Qwen3-Embedding-0.6B & 0.6B & 12.7 & 26.3 & 17.9 & 16.5 & 12.5 & 12.4 & 12.2 & 14.3 & 0.7 & 3.1 & 17.2 & 26.5 & 14.4 \\
Qwen3-Embedding-4B & 4B & 15.9 & 33.6 & 16.7 & 23.2 & 13.2 & 15.0 & 16.8 & 21.2 & 1.7 & 4.0 & 18.4 & 35.4 & 17.9 \\
Qwen3-Embedding-8B & 8B & 14.7 & 17.9 & 15.5 & 19.9 & 9.1 & 12.9 & 16.5 & 17.4 & 0.8 & 2.5 & 16.8 & 24.5 & 14.0 \\
\midrule
\multicolumn{15}{l}{\textit{\textbf{Basic Contrastive Learning}}} \\
Fair Baseline (Qwen3-0.6B) & 0.6B & 28.8 & 31.6 & \underline{25.2} & 28.1 & \underline{15.1} & 22.3 & 24.2 & 8.8 & \textbf{3.5} & \textbf{2.1} & \underline{14.1} & 15.4 & 18.3 \\
Fair Baseline (Qwen3-8B)& 8B & 49.7 & 51.2 & 26.9 & 37.4 & \underline{23.4} & 28.0 & \underline{34.1} & \underline{3.7} & 3.2 & \textbf{2.8} & \underline{16.8} & \underline{31.8} & 25.7 \\
Fair Baseline (LLaMA3.1-8B) & 8B & \underline{57.7} & 40.9 & 24.6 & 29.8 & \textbf{20.1} & \underline{29.2} & 25.8 & \underline{4.3} & \underline{5.7} & \textbf{1.6} & \underline{12.0} & \underline{17.7} & \underline{22.5} \\
\midrule
\multicolumn{15}{l}{\textit{\textbf{Explicit Reasoning}}} \\
Rewrite-then-Retrieve (Qwen3-0.6B) $\dagger$ & 0.6B & 57.8 & 51.6 & 14.1 & 39.4 & 15.0 & 19.8 & 23.7 & 0.6 & 14.6 & 1.2 & 14.9 & 15.5 & 22.4 \\
Rewrite-then-Retrieve (Qwen3-8B) $\dagger$ & 8B & 53.1 & 54.3 & 32.1 & 34.8 & 20.5 & 31.1 & 32.2 & 3.2 & 15.2 & 4.1 & 17.4 & 38.8 & 28.1 \\
Rewrite-then-Retrieve (LLaMA3.1-8B) $\dagger$ & 8B & 63.1 & 54.5 & 27.7 & 40.6 & 17.1 & 28.1 & 28.2 & 1.6 & 11.4 & 3.8 & 16.1 & 30.3 & 26.9 \\
Search-R3 (Qwen2.5-1.5B)& 1.5B & 13.8 & 8.3 & 4.2 & 3.5 & 4.5 & 12.4 & 4.6 & 16.7 & 3.0 & 0.6 & 8.5 & 11.7 & 7.7 \\
\midrule
\multicolumn{15}{l}{\textit{\textbf{Latent Reasoning}}} \\
GIRCSE (Qwen3-0.6B) & 0.6B & \underline{29.0} & \underline{32.8} & 24.7 & \underline{30.6} & 13.6 & \underline{24.0} & \textbf{26.6} & \textbf{11.1} & 1.1 & \underline{1.3} & 13.5 & \underline{20.6} & \underline{19.1} \\
GIRCSE (Qwen3-8B) & 8B & \textbf{59.0} & \textbf{56.5} & \underline{27.2} & \underline{40.3} & 19.0 & \underline{28.5} & 31.4 & 3.2 & \underline{3.6} & \underline{1.7} & 14.0 & 27.2 & \underline{26.0} \\
GIRCSE (LLaMA3.1-8B) & 8B & 50.6 & \underline{43.4} & \textbf{28.3} & \underline{35.7} & 14.3 & 26.9 & \underline{26.4} & \textbf{6.0} & 5.2 & 0.5 & 11.6 & 15.1 & 22.0 \\
\rowcolor[RGB]{236,244,252} \textbf{\ours{}} (Qwen3-0.6B) & 0.6B & \textbf{50.0} & \textbf{45.9} & \textbf{25.7} & \textbf{32.4} & \textbf{18.4} & \textbf{27.1} & \underline{26.5} & \underline{9.1} & \underline{2.7} & 1.2 & \textbf{16.0} & \textbf{22.4} & \textbf{23.1} \\
\rowcolor[RGB]{236,244,252} \textbf{\ours{}} (Qwen3-8B) & 8B & \underline{58.0} & \underline{51.8} & \textbf{29.0} & \textbf{44.1} & \textbf{26.0} & \textbf{32.9} & \textbf{34.5} & \textbf{9.2} & \textbf{11.7} & 1.6 & \textbf{18.7} & \textbf{34.0} & \textbf{29.3} \\
\rowcolor[RGB]{236,244,252} \textbf{\ours{}} (LLaMA3.1-8B) & 8B & \textbf{58.4} & \textbf{48.1} & \underline{28.0} & \textbf{40.9} & \underline{17.0} & \textbf{29.9} & \textbf{28.3} & 1.7 & \textbf{5.9} & \underline{1.5} & \textbf{14.6} & \textbf{19.2} & \textbf{24.4} \\
\bottomrule
\end{tabular}
}
\end{table*}

\header{Standard Dense Retrievers:} We include state-of-the-art encoder-based models such as BGE-M3~\cite{bge_m3} and E5-Large-Instruct~\cite{e5}, as well as LLM-based retrievers including E5-Mistral~\cite{e5mistral} and Qwen3-Embedding~\cite{qwen3embedding}. Additionally, we train a ``Fair Baseline'' using standard contrastive learning on our composite dataset to isolate the gains attributed to our architecture.
    
\header{Pipeline Methods (Rewrite-then-Retrieve):} This category employs an external LLM to explicitly rewrite the query before retrieval. We utilize BGE-Reasoner-Rewriter-7B~\cite{reasonembed} to generate rewritten queries as input to the retriever (using the Fair Baseline model) during inference.
    
\header{Explicit Reasoning Methods:} These approaches integrate generation and retrieval within a single model, generating an explicit CoT prior to outputting the representation. We compare against Search-R3~\cite{search-r3} and GRACE~\cite{grace}. As training codes for some baselines are not publicly available, we perform inference using their official checkpoints.
    
\header{Implicit Reasoning Methods:} We compare against GIRCSE, a representative method that also employs latent tokens to model reasoning steps before producing the final embedding.

\subsection{Implementation Details}
\label{sec:implementation}
We conduct experiments using base version of Qwen3 series (0.6B, 4B, 8B) and LLaMA 3.2 series (1B, 3B, 8B). As LLaMA 3.2 does not offer an 8B variant, we utilize LLaMA 3.1-8B to evaluate performance at that scale. 
All models are fine-tuned for 1 epoch using LoRA ($r=64, \alpha=32$) on 4 A100 GPUs, with a global batch size of 8 (batch size per device set to 2 with 8 gradient accumulation steps). We use the AdamW optimizer with a learning rate of 1e-4 and a warmup ratio of 0.1. Each query is paired with one hard negative sample and in-batch negatives for training. The maximum sequence length for both query and documents is set to 512 during training and expanded to 8192 during testing to accommodate longer contexts. To accommodate the explicit reasoning paths processed by the Explicit-View, we set the maximum sequence length for this view to 1024 during training.  
The temperature hyperparameters for both the contrastive loss ($\tau$) and the distillation loss ($\tau_{kd}$) are set to 0.02. To balance the multi-task objective, we set the loss weights as follows: $\lambda_1=1$, $\lambda_2=10$, and $\lambda_3=0.1$. The number of latent thinking steps $K$ is set to 3 for both training and inference phases.

\section{Results}
In this section, we present the experimental results of \ours{}, followed by a detailed ablation study and efficiency analysis. Our evaluation aims to answer the following three research questions:

RQ1: How does our proposed framework perform compared to fair baselines utilizing standard contrastive learning and other training methods?

RQ2: What is the impact of the proposed multi-grained self-distillation strategies? 

RQ3: Can \ours{} achieve a superior trade-off between retrieval performance and  latency compared to other baseline methods?

\subsection{Main Results}
In this section, we present a comprehensive analysis of \ours{}'s performance on the in-domain reasoning benchmark (BRIGHT) and out-of-domain benchmarks (FollowIR and BrowseComp-Plus). To ensure a fair comparison, we reproduced fair baselines using identical training data across all methods and conducted experiments using three model scales based on two distinct architectures. The main results are reported in Table~\ref{tab:main_results} and Table~\ref{tab:main_results_2} (experiments on the full range of parameter scales are detailed in Subsection~\ref{sec:robust_analysis}). Our key observations are summarized below.

\header{\ours{} Shows Significant Improvements over Standard Dense Retrievers.} 
First, \ours{} demonstrates a significant improvement over fair baselines trained with standard contrastive learning. As shown in Table~\ref{tab:main_results}, \ours{} (Qwen3-8B) achieves an average nDCG@10 of 29.3. This performance not only surpasses the original Qwen3-Embedding-8B (14.0) but also exceeds the fine-tuned Fair Baseline (25.7) by approximately 15\%. This substantial performance gap indicates that standard contrastive training with conventional retriever architectures is insufficient for complex reasoning tasks, as it fails to leverage the model's underlying reasoning potential. By internalizing reasoning into the latent space, \ours{} effectively activates the reasoning capabilities of the LLM backbone, bridging the gap between shallow semantic matching and complex reasoning.

\header{\ours{} Matches the Upper Bound of of Explicit Pipelines.} 
A key motivation of our work is to match the performance of computationally heavy ``rewrite-then-retrieve'' pipelines without incurring their high latency costs. Experimental results demonstrate that on both Qwen3-0.6B and Qwen3-8B backbones, our method outperforms explicit reasoning approaches based on query rewriting (surpassing them by 0.7 and 1.2 points, respectively). Furthermore, \ours{} significantly outperforms explicit reasoning models that require generating multiple thinking tokens, such as Search-R3 and InBedder. Crucially, \ours{} achieves this competitive performance using only a few latent tokens, thereby avoiding the prohibitive computational cost associated with decoding long explicit CoT rationales. This validates our core insight: the semantics of reasoning paths can be effectively compressed into latent states via self-distillation. With an appropriate backbone, our method more effectively elicits the model's inherent reasoning capabilities, leading to superior performance.

\header{\ours{} Superiority over Existing Implicit Reasoning Methods.} 
We further compare \ours{} with state-of-the-art implicit reasoning methods, specifically GIRCSE. Across 9 experimental settings involving different model scales and datasets, \ours{} outperforms GIRCSE in 8 cases. Given that both methods utilize identical input contexts and computational budgets, these results highlight the effectiveness of our proposed self-distillation framework. This suggests that integrating supervision from the teacher model significantly aids the learning of latent thinking. Unlike GIRCSE, which relies solely on output contrastive signals, \ours{} employs a dual alignment strategy which incorporate both process-level and output-level distillation to ensure that intermediate latent tokens capture richer semantic information.

\begin{table*}[t]
\centering
\caption{Retrieval performance. Columns under Robust04, News21, and Core17 belong to the \textbf{FollowIR} benchmark.}
\label{tab:main_results_2}
\resizebox{\textwidth}{!}{
\begin{tabular}{l|c|cc|cc|cc|cc|ccc}
\toprule
\multirow{2}{*}{\textbf{Models}} & \multirow{2}{*}{\textbf{Size}} & \multicolumn{2}{c|}{\textbf{Robust04}} & \multicolumn{2}{c|}{\textbf{News21}} & \multicolumn{2}{c|}{\textbf{Core17}} & \multicolumn{2}{c|}{\textbf{FollowIR-Avg}} & \multicolumn{3}{c}{\textbf{BrowseComp-Plus}} \\
\cmidrule(lr){3-4} \cmidrule(lr){5-6} \cmidrule(lr){7-8} \cmidrule(lr){9-10} \cmidrule(lr){11-13}
& & MAP@5 & p-MRR & nDCG@5 & p-MRR & MAP@5 & p-MRR & Score & p-MRR &  R@5 & R@100 & R@1000  \\
\midrule
\multicolumn{13}{l}{\textit{\textbf{Standard Dense Retrievers}}} \\
BGE-M3 & 0.6B & 1.5 & -1.6 & 21.4 & -1.3 & 7.5 & -8.8 & 10.1 & -3.9 & 4.1 & 21.7 & 47.2 \\ 
Multilingual-E5-Large-Instruct & 0.6B & 2.4 & 2.7 & 25.0 & 1.2 & 8.5 & -6.0 & 12.0 & -0.7 & 10.1 & 36.8 & 68.6 \\ 
E5-Mistral-7B-Instruct & 7B & 2.7 & 2.5 & 28.8 & 0.8 & 13.7 & -7.8 & 15.1 & -1.5 & 9.3 & 37.1 & 70.2 \\ 
Qwen3-Embedding-0.6B & 0.6B & 2.8 & 8.9 & 27.3 & 3.6 & 11.4 & 2.7 & 13.8 & 5.1 & 8.0 & 29.1 & 64.8 \\ 
Qwen3-Embedding-4B & 4B & 5.0 & 11.0 & 26.7 & 6.9 & 11.2 & 11.1 & 14.3 & 9.7 & 5.9 & 18.1 & 39.4 \\ 
Qwen3-Embedding-8B & 8B & 3.1 & 6.2 & 25.5 & 8.8 & 10.1 & 6.6 & 12.9 & 7.2 & 7.7 & 31.6 & 61.3 \\ 
\midrule
\multicolumn{13}{l}{\textit{\textbf{Basic Contrastive Learning}}} \\
Fair Baseline (Qwen3-0.6B) & 0.6B & \underline{2.1} & \underline{3.5} & \underline{13.4} & \underline{-0.6} & \underline{6.7} & \textbf{-3.4} & \underline{7.4} & \underline{-0.2} & 3.5 & 21.3 & 46.9 \\
Fair Baseline (Qwen3-8B) & 8B & 2.8 & \underline{4.4} & 18.9 & \textbf{2.0} & \underline{11.2} & \textbf{-1.3} & 11.0 & 1.7 & 11.3 & 37.4 & 63.2 \\
Fair Baseline (LLaMA3.1-8B) & 8B & 2.5 & \textbf{2.8} & 18.9 & \underline{0.1} & 8.1 & -2.6 & 9.8 & \underline{0.1} & 6.1 & \underline{25.4} & 50.8 \\
\midrule
\multicolumn{13}{l}{\textit{\textbf{Explicit Reasoning}}} \\
Search-R3 (Qwen2.5-1.5B) & 1.5B & 3.2 & 7.1 & 26.2 & -0.1 & 10.1 & 2.7 & 13.2 & 3.2 & 0.0 & 0.3 & 1.1 \\
Inbedder (LLaMA2-7B) & 7B & 2.6 & 3.2 & 8.5 & 3.1 & 1.6 & -6.4 & 4.2 & -0.1 & 1.4 & 8.1 & 27.4 \\
\midrule
\multicolumn{13}{l}{\textit{\textbf{Latent Reasoning}}} \\
GIRCSE (Qwen3-0.6B) & 0.6B & \textbf{3.0} & \textbf{3.9} & 12.8 & -0.9 & 5.7 & -7.7 & 7.2 & -1.6 & \textbf{6.9} & \underline{25.2} & \underline{52.8} \\
GIRCSE (Qwen3-8B) & 8B & \underline{3.0} & 4.2 & \textbf{22.6} & \underline{0.7} & 8.5 & \underline{1.0} & \underline{11.4} & \underline{2.0} & \textbf{13.0} & \textbf{40.8} & \textbf{68.1} \\
GIRCSE (LLaMA3.1-8B) & 8B & \underline{2.9} & 1.0 & \underline{19.3} & -1.3 & \textbf{12.2} & \textbf{-1.2} & \underline{11.5} & -0.5 & \underline{6.7} & 23.5 & \underline{51.0} \\
\rowcolor[RGB]{236,244,252} \textbf{\ours{}} (Qwen3-0.6B) & 0.6B & 2.0 & 3.4 & \textbf{25.0} & \textbf{1.0} & \textbf{7.4} & \underline{-4.5} & \textbf{11.5} & \textbf{0.0} & \underline{6.8} & \textbf{26.8} & \textbf{54.9} \\
\rowcolor[RGB]{236,244,252} \textbf{\ours{}} (Qwen3-8B) & 8B & \textbf{4.1} & \textbf{5.8} & \underline{21.8} & 0.2 & \textbf{11.4} & 1.3 & \textbf{12.5} & \textbf{2.4} & \underline{11.7} & \underline{38.4} & \underline{66.9} \\
\rowcolor[RGB]{236,244,252} \textbf{\ours{}} (LLaMA3.1-8B) & 8B & \textbf{3.6} & \underline{2.7} & \textbf{22.0} & \textbf{1.6} & \underline{11.1} & \underline{-1.9} & \textbf{12.2} & \textbf{0.8} & \textbf{6.8} & \textbf{25.7} & \textbf{52.9} \\
\bottomrule
\end{tabular}
}
\end{table*}


\subsection{Ablation Study}
To address RQ2, we conduct a comprehensive ablation study utilizing Qwen3-0.6B as the backbone model. Given that \ours{} functions as a self-distillation framework designed to internalize explicit reasoning, our analysis focuses on two critical dimensions:

\header{(1) Architectural Effectiveness.} We examine the contribution of the dual-view structure and the latent thinking mechanism. \textbf{w/o Explicit View} removes the explicit-view branch entirely, training the student to perform latent reasoning relying solely on contrastive loss without privileged supervision. \textbf{w/o Latent View} replaces the latent reasoning mechanism with a standard single-forward pass to obtain the final embedding, assessing whether the shared backbone learning benefits standard architectures even in the absence of latent tokens.

\begin{table}[t]
    \centering
    \small
    \caption{Ablation study of our proposed method. For FollowIR, we report the average score on all subsets.}
    \label{tab:ablation}
    \setlength{\tabcolsep}{0.7mm}{}  
    \resizebox{\linewidth}{!}{
        \begin{tabular}{l c cc cc}
        \toprule
        \multirow{2}{*}{\textbf{Method}} & \textbf{Bright} & \multicolumn{2}{c}{\textbf{FollowIR}} & \multicolumn{2}{c}{\textbf{BrowseComp-Plus}} \\
        \cmidrule(lr){2-2} \cmidrule(lr){3-4} \cmidrule(lr){5-6}
        
         & nDCG@10 & Score & p-MRR & R@5 & R@1000 \\
        \midrule
        
        \rowcolor[RGB]{236,244,252} \textbf{\ours{} (Qwen3-0.6B)} & \textbf{23.10} & \underline{11.49} & \underline{-0.02} & \underline{6.76} & \textbf{54.88} \\
        \midrule
\quad        w/o Explicit View & 20.59 & 9.78 & -1.71 & 6.48 & 50.14 \\
\quad        w/o Latent View & 19.93 & 8.62 & -0.11 & 5.93 & 52.25 \\
        \midrule
\quad        w/o Process Align. & \underline{22.33} & \textbf{11.63} & -1.58 & 6.18 & 52.43 \\
\quad        w/o Output Distill. & 19.97 & 7.55 & \textbf{0.43} & \textbf{7.03} & \underline{53.02} \\
\quad        w/o Co-Learning & 20.98 & 10.87 & -0.54 & 5.56 & 52.57 \\
        \midrule
\quad        Basic Contrastive Learn. & 18.25 & 7.38 & -0.17 & 3.54 & 46.9 \\
        
        \bottomrule
        \end{tabular}
    }
\end{table}

\header{(2) Learning Strategy Effectiveness.} We isolate the impact of the co-learning paradigm and our alignment objectives. \textbf{w/o Process KD} excludes the trajectory alignment loss ($\mathcal{L}_{kl}^{mid}$), isolating the impact of supervising intermediate latent states. \textbf{w/o Output-level Distillation} removes the final output alignment loss ($\mathcal{L}_{kl}^{out}$), retaining only process-level supervision. \textbf{w/o Co-Learning} replaces online self-distillation with a two-stage process, where a fixed explicit-view teacher (fine-tuned offline) generates static supervision labels, validating the efficacy of our dynamic shared-backbone training. 

\begin{figure}[!t]
    \centering
    \includegraphics[width=0.9\linewidth]{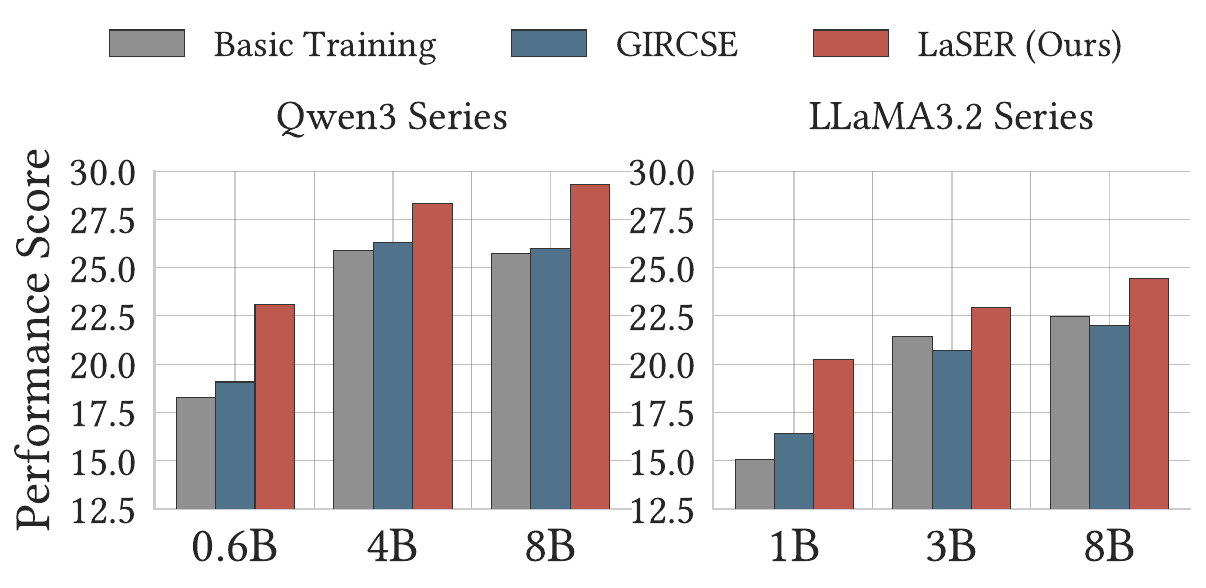}
    \caption{
    Performance comparison of \ours{} and baselines across different backbones and model sizes on Bright.  
    }
    \label{fig:robust_analysis}
\end{figure}

\header{Architectural Effectiveness.} We first validate the necessity of the proposed latent reasoning mechanism. As shown in Table~\ref{tab:ablation}, removing the latent view leads to the most significant performance degradation (e.g., from 23.10 to 19.93 on Bright). This suggests that the single-vector output form inherent to standard dense retrievers creates a representational bottleneck that constrains reasoning capabilities and hinders the model from effectively absorbing the additional information provided by the teacher input. In contrast, our latent thinking tokens successfully expand the semantic capacity. 
We also examine the role of the explicit teacher. Removing the explicit view similarly impairs performance. This indicates that without explicit CoT supervision, the learning efficacy of the intermediate latent tokens diminishes, likely due to the absence of explicit semantic guidance.

\header{Impact of Multi-grained Alignment.} We further analyze the effectiveness of our learning strategies. Experimental results demonstrate that both output-level and process-level distillation are essential; removing either component significantly compromises reasoning capabilities and generalization performance across tasks. Notably, the exclusion of output distillation leads to a more pronounced decline, confirming that transferring fine-grained ranking preferences from the teacher is a prerequisite for effective student learning. We also observe that online learning has a substantial impact. On both in-domain and out-of-domain datasets, allowing the teacher to learn dynamically yields superior results. This validates that our shared backbone design promotes better mutual adaptation between views (further analysis of this claim is provided in Section~\ref{sec:both_view}).

\subsection{Robust Analysis}
\label{sec:robust_analysis}

We verify the robustness of \ours{} across different backbone architectures, key observations are two fold:

\header{Consistent Superiority over Baselines.} We conduct experiments with various backbone model on Bright. As shown in Figure~\ref{fig:robust_analysis}, \ours{} maintains consistent superiority over all baselines across diverse model families in all 6 settings. Notably, on lightweight backbones (e.g., 0.6B and 1B), \ours{} demonstrates superior effectiveness (23.10), even surpassing the computationally intensive rewrite pipeline (22.35). This indicates that our self-distillation framework enables small models to ``think'' effectively without the overhead of external reasoners.

\header{Instability of Contrastive-only Latent Thinking.} We observe that GIRCSE, which relies solely on the contrastive learning objective, exhibits instability across different architectures. In certain settings (e.g., LLaMA3.2-3B), it yields performance degradations compared to the standard retriever baseline, while showing only marginal improvements in others. In contrast, by incorporating distillation signals, our method achieves substantial gains on similar architectures. This suggests that latent thinking mechanisms may require privileged supervision to facilitate effective learning.


\begin{figure}[!t]
    \centering
    \includegraphics[width=0.95\linewidth]{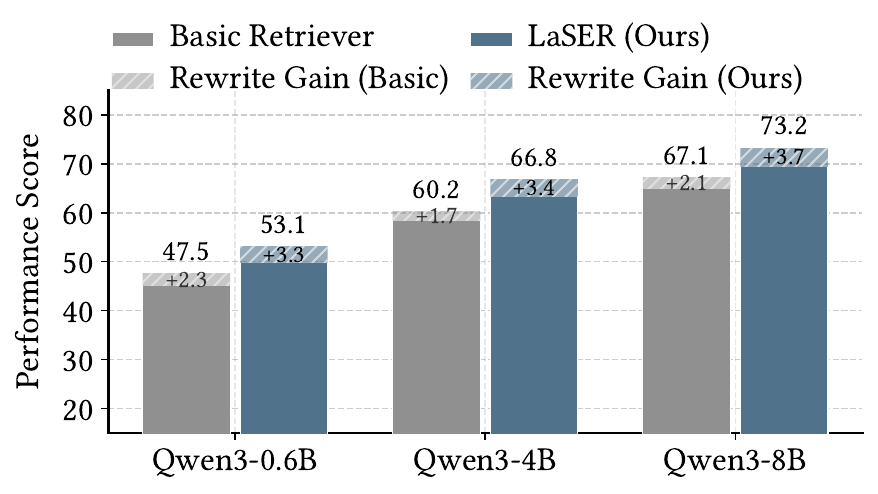}
    \caption{
    Comparison of the performance of using rewrite path on Bright between our method and basic training.
    }
    \label{fig:rewrite_comp}
\end{figure}

\subsection{Latency Analysis}
To address RQ3, we evaluate inference latency on a subset of 80 queries from the BRIGHT benchmark, utilizing a single NVIDIA A100 (80GB) GPU. 
To ensure a rigorous comparison, we employ vLLM~\cite{vllm} for the inference of external rewriter models in explicit reasoning pipelines, while the native Hugging Face Transformers implementation is used for all retriever backbones (including ours). 
A batch size of 8 is maintained for all baselines and our method.
As demonstrated in Figure~\ref{fig:intro_fig}, \ours{} drastically reduces computational costs, incurring only 0.3\% of the latency associated with the ``rewrite-then-retrieve'' pipeline while maintaining comparable retrieval effectiveness. 
Although \ours{} introduces a modest latency overhead compared to the standard base retriever (approximately $1.7\times$) due to the autoregressive generation of latent thinking tokens, this overhead decreases as model scale increases.  For larger backbones, the relative cost of vector operations becomes marginal compared to the model's forward pass time, allowing the system to benefit more significantly from KV caching mechanisms.

\subsection{Impact of Latent Reasoning Steps}
\begin{figure}[!t]
    \centering
    \includegraphics[width=\linewidth]{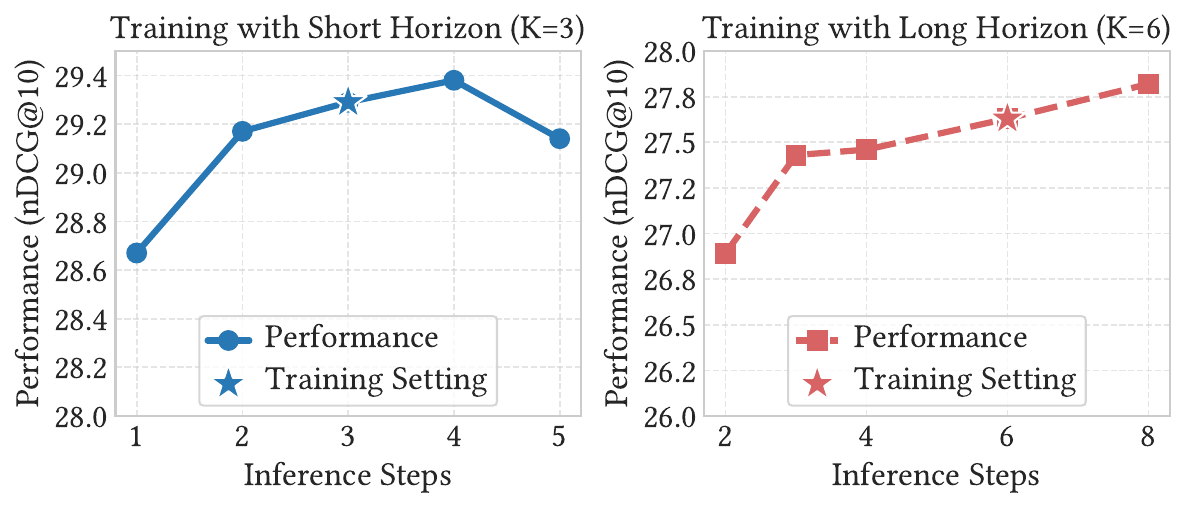}
    \caption{
    Effect of latent thinking steps during training and inference.
    }
    \label{fig:step_analysis}
\end{figure}

We investigate the impact of the number of latent thinking steps ($K$) during both the training and inference on Qwen3-8B, with results illustrated in Figure~\ref{fig:step_analysis}.

\header{Training Efficiency.} We observe that increasing the training steps from $K=3$ to $K=6$ yields negligible performance gains or even slight degradation. We attribute this phenomenon to the high semantic density of the latent space facilitated by our framework. Unlike standard implicit reasoning, the introduction of the Explicit-View provides privileged supervision, serving as a clear optimization target. This allows \ours{} to effectively compress the reasoning process, where a few latent steps are sufficient to capture the core logic of the reasoning path. Consequently, enforcing a larger $K$ during training may introduce noise by compelling the model to align with granular, non-informative segments of the explicit text.

\header{Inference Scaling.} Conversely, increasing $K$ during inference demonstrates consistent performance improvements, which highlights that \ours{} has acquired a robust mechanism for iterative refinement of embedding. Rather than merely memorizing fixed step trajectories or copying preceding states, the model actively utilizes additional computing steps to deepen its semantic understanding and optimize the query representation.

\subsection{Analysis of Co-Learning of Both Views}
\label{sec:both_view}

Since our framework involves the joint training of two distinct views sharing a common backbone, we further investigate the synergistic interaction between them. Specifically, we evaluate the backbone's capability to process explicit reasoning by comparing the performance gains achieved when providing explicit rewritten queries to both the Basic Retriever and \ours{} during inference.
As shown in Figure~\ref{fig:rewrite_comp}, \ours{} demonstrates a significantly larger performance boost from explicit rewrites compared to the Basic baseline across all model sizes (e.g. 3.4 v.s. 1.7 in Qwen3-4B). This substantial difference indicates that our Explicit-View training objective effectively optimizes the shared backbone, enabling it to better capture and encode the semantic information within CoT rationales for retrieval tasks. Consequently, this enhanced explicit processing capability serves as a superior semantic upper bound, thereby facilitating more effective distillation into the Latent-View.

\section{Conclusion \& Future Work}

In this paper, we propose \ours{}, a novel self-distillation framework that internalizes explicit CoT reasoning into the latent space of dense retrievers. By employing a multi-grained alignment strategy that synchronizes latent thinking tokens with explicit reasoning trajectories, \ours{} effectively bridges the gap between deep semantic reasoning and retrieval efficiency. Extensive experiments demonstrate that \ours{} significantly outperforms state-of-the-art baselines, achieving performance comparable to computationally expensive ``rewrite-then-retrieve'' pipelines while maintaining low inference latency. Our work validates that the semantics of explicit reasoning can be effectively compressed into latent states, offering an efficient solution for complex query understanding. 
Despite these promising results,  our method represents only a preliminary step toward fully autonomous latent reasoning. In future work, we plan to explore reinforcement learning techniques to further optimize the intermediate reasoning process by directly optimizing the latent reasoning trajectory based on retrieval utility.


\bibliographystyle{ACM-Reference-Format}
\bibliography{reference}

@inproceedings{query2doc,
  author       = {Liang Wang and
                  Nan Yang and
                  Furu Wei},
  editor       = {Houda Bouamor and
                  Juan Pino and
                  Kalika Bali},
  title        = {Query2doc: Query Expansion with Large Language Models},
  booktitle    = {Proceedings of the 2023 Conference on Empirical Methods in Natural
                  Language Processing, {EMNLP} 2023, Singapore, December 6-10, 2023},
  pages        = {9414--9423},
  publisher    = {Association for Computational Linguistics},
  year         = {2023},
  url          = {https://doi.org/10.18653/v1/2023.emnlp-main.585},
  doi          = {10.18653/V1/2023.EMNLP-MAIN.585},
  bibsource    = {dblp computer science bibliography, https://dblp.org}
}

@article{BrowseComp-plus,
  author       = {Zijian Chen and
                  Xueguang Ma and
                  Shengyao Zhuang and
                  Ping Nie and
                  Kai Zou and
                  Andrew Liu and
                  Joshua Green and
                  Kshama Patel and
                  Ruoxi Meng and
                  Mingyi Su and
                  Sahel Sharifymoghaddam and
                  Yanxi Li and
                  Haoran Hong and
                  Xinyu Shi and
                  Xuye Liu and
                  Nandan Thakur and
                  Crystina Zhang and
                  Luyu Gao and
                  Wenhu Chen and
                  Jimmy Lin},
  title        = {BrowseComp-Plus: {A} More Fair and Transparent Evaluation Benchmark
                  of Deep-Research Agent},
  journal      = {CoRR},
  volume       = {abs/2508.06600},
  year         = {2025},
  url          = {https://doi.org/10.48550/arXiv.2508.06600},
  doi          = {10.48550/ARXIV.2508.06600},
  eprinttype    = {arXiv},
  eprint       = {2508.06600},
  bibsource    = {dblp computer science bibliography, https://dblp.org}
}

@article{latent_cot_survey,
  author       = {Xinghao Chen and
                  Anhao Zhao and
                  Heming Xia and
                  Xuan Lu and
                  Hanlin Wang and
                  Yanjun Chen and
                  Wei Zhang and
                  Jian Wang and
                  Wenjie Li and
                  Xiaoyu Shen},
  title        = {Reasoning Beyond Language: {A} Comprehensive Survey on Latent Chain-of-Thought
                  Reasoning},
  journal      = {CoRR},
  volume       = {abs/2505.16782},
  year         = {2025},
  url          = {https://doi.org/10.48550/arXiv.2505.16782},
  doi          = {10.48550/ARXIV.2505.16782},
  eprinttype    = {arXiv},
  eprint       = {2505.16782},
  bibsource    = {dblp computer science bibliography, https://dblp.org}
}

@article{adaptive_distillation,
  author       = {Zhengyang Liang and
                  Meiyu Liang and
                  Wei Huang and
                  Yawen Li and
                  Zhe Xue},
  title        = {Dynamic Self-adaptive Multiscale Distillation from Pre-trained Multimodal
                  Large Model for Efficient Cross-modal Representation Learning},
  journal      = {CoRR},
  volume       = {abs/2404.10838},
  year         = {2024},
  url          = {https://doi.org/10.48550/arXiv.2404.10838},
  doi          = {10.48550/ARXIV.2404.10838},
  eprinttype    = {arXiv},
  eprint       = {2404.10838},
  bibsource    = {dblp computer science bibliography, https://dblp.org}
}

@article{codi,
  author       = {Zhenyi Shen and
                  Hanqi Yan and
                  Linhai Zhang and
                  Zhanghao Hu and
                  Yali Du and
                  Yulan He},
  title        = {{CODI:} Compressing Chain-of-Thought into Continuous Space via Self-Distillation},
  journal      = {CoRR},
  volume       = {abs/2502.21074},
  year         = {2025},
  url          = {https://doi.org/10.48550/arXiv.2502.21074},
  doi          = {10.48550/ARXIV.2502.21074},
  eprinttype    = {arXiv},
  eprint       = {2502.21074},
  bibsource    = {dblp computer science bibliography, https://dblp.org}
}

@article{distillation_sota,
  author       = {Dun Zhang and
                  FulongWang},
  title        = {Jasper and Stella: distillation of {SOTA} embedding models},
  journal      = {CoRR},
  volume       = {abs/2412.19048},
  year         = {2024},
  url          = {https://doi.org/10.48550/arXiv.2412.19048},
  doi          = {10.48550/ARXIV.2412.19048},
  eprinttype    = {arXiv},
  eprint       = {2412.19048},
  bibsource    = {dblp computer science bibliography, https://dblp.org}
}

@article{coconut,
  author       = {Shibo Hao and
                  Sainbayar Sukhbaatar and
                  DiJia Su and
                  Xian Li and
                  Zhiting Hu and
                  Jason Weston and
                  Yuandong Tian},
  title        = {Training Large Language Models to Reason in a Continuous Latent Space},
  journal      = {CoRR},
  volume       = {abs/2412.06769},
  year         = {2024},
  url          = {https://doi.org/10.48550/arXiv.2412.06769},
  doi          = {10.48550/ARXIV.2412.06769},
  eprinttype    = {arXiv},
  eprint       = {2412.06769},
  bibsource    = {dblp computer science bibliography, https://dblp.org}
}

@article{knowledge_distillation_survey,
  author       = {Amir M. Mansourian and
                  Rozhan Ahmadi and
                  Masoud Ghafouri and
                  Amir Mohammad Babaei and
                  Elaheh Badali Golezani and
                  Zeynab Yasamani Ghamchi and
                  Vida Ramezanian and
                  Alireza Taherian and
                  Kimia Dinashi and
                  Amirali Miri and
                  Shohreh Kasaei},
  title        = {A Comprehensive Survey on Knowledge Distillation},
  journal      = {Trans. Mach. Learn. Res.},
  volume       = {2025},
  year         = {2025},
  url          = {https://openreview.net/forum?id=3cbJzdR78B},
  bibsource    = {dblp computer science bibliography, https://dblp.org}
}

@inproceedings{inbedder,
  author       = {Letian Peng and
                  Yuwei Zhang and
                  Zilong Wang and
                  Jayanth Srinivasa and
                  Gaowen Liu and
                  Zihan Wang and
                  Jingbo Shang},
  editor       = {Lun{-}Wei Ku and
                  Andre Martins and
                  Vivek Srikumar},
  title        = {Answer is All You Need: Instruction-following Text Embedding via Answering
                  the Question},
  booktitle    = {Proceedings of the 62nd Annual Meeting of the Association for Computational
                  Linguistics (Volume 1: Long Papers), {ACL} 2024, Bangkok, Thailand,
                  August 11-16, 2024},
  pages        = {459--477},
  publisher    = {Association for Computational Linguistics},
  year         = {2024},
  url          = {https://doi.org/10.18653/v1/2024.acl-long.27},
  doi          = {10.18653/V1/2024.ACL-LONG.27},
  bibsource    = {dblp computer science bibliography, https://dblp.org}
}

@article{rader,
  author       = {Debrup Das and
                  Se{\'{a}}n {\'{O}} Nuall{\'{a}}in and
                  Razieh Rahimi},
  title        = {RaDeR: Reasoning-aware Dense Retrieval Models},
  journal      = {CoRR},
  volume       = {abs/2505.18405},
  year         = {2025},
  url          = {https://doi.org/10.48550/arXiv.2505.18405},
  doi          = {10.48550/ARXIV.2505.18405},
  eprinttype    = {arXiv},
  eprint       = {2505.18405},
  bibsource    = {dblp computer science bibliography, https://dblp.org}
}

@article{reasonrank,
  author       = {Wenhan Liu and
                  Xinyu Ma and
                  Weiwei Sun and
                  Yutao Zhu and
                  Yuchen Li and
                  Dawei Yin and
                  Zhicheng Dou},
  title        = {ReasonRank: Empowering Passage Ranking with Strong Reasoning Ability},
  journal      = {CoRR},
  volume       = {abs/2508.07050},
  year         = {2025},
  url          = {https://doi.org/10.48550/arXiv.2508.07050},
  doi          = {10.48550/ARXIV.2508.07050},
  eprinttype    = {arXiv},
  eprint       = {2508.07050},
  bibsource    = {dblp computer science bibliography, https://dblp.org}
}

@article{tongsearch,
  author       = {Xubo Qin and
                  Jun Bai and
                  Jiaqi Li and
                  Zixia Jia and
                  Zilong Zheng},
  title        = {TongSearch-QR: Reinforced Query Reasoning for Retrieval},
  journal      = {CoRR},
  volume       = {abs/2506.11603},
  year         = {2025},
  url          = {https://doi.org/10.48550/arXiv.2506.11603},
  doi          = {10.48550/ARXIV.2506.11603},
  eprinttype    = {arXiv},
  eprint       = {2506.11603},
  bibsource    = {dblp computer science bibliography, https://dblp.org}
}

@article{reasonembed,
  author       = {Jianlyu Chen and
                  Junwei Lan and
                  Chaofan Li and
                  Defu Lian and
                  Zheng Liu},
  title        = {ReasonEmbed: Enhanced Text Embeddings for Reasoning-Intensive Document
                  Retrieval},
  journal      = {CoRR},
  volume       = {abs/2510.08252},
  year         = {2025},
  url          = {https://doi.org/10.48550/arXiv.2510.08252},
  doi          = {10.48550/ARXIV.2510.08252},
  eprinttype    = {arXiv},
  eprint       = {2510.08252},
  bibsource    = {dblp computer science bibliography, https://dblp.org}
}

@article{search-r3,
  author       = {Yuntao Gui and
                  James Cheng},
  title        = {Search-R3: Unifying Reasoning and Embedding Generation in Large Language
                  Models},
  journal      = {CoRR},
  volume       = {abs/2510.07048},
  year         = {2025},
  url          = {https://doi.org/10.48550/arXiv.2510.07048},
  doi          = {10.48550/ARXIV.2510.07048},
  eprinttype    = {arXiv},
  eprint       = {2510.07048},
  bibsource    = {dblp computer science bibliography, https://dblp.org}
}

@article{think_then_embed,
  author       = {Xuanming Cui and
                  Jianpeng Cheng and
                  Hong{-}you Chen and
                  Satya Narayan Shukla and
                  Abhijeet Awasthi and
                  Xichen Pan and
                  Chaitanya Ahuja and
                  Shlok Kumar Mishra and
                  Yonghuan Yang and
                  Jun Xiao and
                  Qi Guo and
                  Ser{-}Nam Lim and
                  Aashu Singh and
                  Xiangjun Fan},
  title        = {Think Then Embed: Generative Context Improves Multimodal Embedding},
  journal      = {CoRR},
  volume       = {abs/2510.05014},
  year         = {2025},
  url          = {https://doi.org/10.48550/arXiv.2510.05014},
  doi          = {10.48550/ARXIV.2510.05014},
  eprinttype    = {arXiv},
  eprint       = {2510.05014},
  bibsource    = {dblp computer science bibliography, https://dblp.org}
}

@article{diverse_multiquery_retrieval,
  author       = {Hung{-}Ting Chen and
                  Xiang Liu and
                  Shauli Ravfogel and
                  Eunsol Choi},
  title        = {Beyond Single Embeddings: Capturing Diverse Targets with Multi-Query
                  Retrieval},
  journal      = {CoRR},
  volume       = {abs/2511.02770},
  year         = {2025},
  url          = {https://doi.org/10.48550/arXiv.2511.02770},
  doi          = {10.48550/ARXIV.2511.02770},
  eprinttype    = {arXiv},
  eprint       = {2511.02770},
  bibsource    = {dblp computer science bibliography, https://dblp.org}
}

@article{zhang_secretlyreasoner,
  author       = {Yichi Zhang and
                  Jun Bai and
                  Zhixin Cai and
                  Shuhan Qin and
                  Zhuofan Chen and
                  Jinghua Guan and
                  Wenge Rong},
  title        = {Your Dense Retriever is Secretly an Expeditious Reasoner},
  journal      = {CoRR},
  volume       = {abs/2510.21727},
  year         = {2025},
  url          = {https://doi.org/10.48550/arXiv.2510.21727},
  doi          = {10.48550/ARXIV.2510.21727},
  eprinttype    = {arXiv},
  eprint       = {2510.21727},
  bibsource    = {dblp computer science bibliography, https://dblp.org}
}

@article{grace,
  author       = {Jiashuo Sun and
                  Shixuan Liu and
                  Zhaochen Su and
                  Xianrui Zhong and
                  Pengcheng Jiang and
                  Bowen Jin and
                  Peiran Li and
                  Weijia Shi and
                  Jiawei Han},
  title        = {{GRACE:} Generative Representation Learning via Contrastive Policy
                  Optimization},
  journal      = {CoRR},
  volume       = {abs/2510.04506},
  year         = {2025},
  url          = {https://doi.org/10.48550/arXiv.2510.04506},
  doi          = {10.48550/ARXIV.2510.04506},
  eprinttype    = {arXiv},
  eprint       = {2510.04506},
  bibsource    = {dblp computer science bibliography, https://dblp.org}
}

@inproceedings{expandr,
    title = "{E}xpand{R}: Teaching Dense Retrievers Beyond Queries with {LLM} Guidance",
    author = "Yao, Sijia  and
      Huang, Pengcheng  and
      Liu, Zhenghao  and
      Gu, Yu  and
      Yan, Yukun  and
      Yu, Shi  and
      Yu, Ge",
    editor = "Christodoulopoulos, Christos  and
      Chakraborty, Tanmoy  and
      Rose, Carolyn  and
      Peng, Violet",
    booktitle = "Proceedings of the 2025 Conference on Empirical Methods in Natural Language Processing",
    month = nov,
    year = "2025",
    address = "Suzhou, China",
    publisher = "Association for Computational Linguistics",
    url = "https://aclanthology.org/2025.emnlp-main.963/",
    doi = "10.18653/v1/2025.emnlp-main.963",
    pages = "19036--19054",
    ISBN = "979-8-89176-332-6"
}

@inproceedings{conv_search_rewrite,
    title = "Learning Contextual Retrieval for Robust Conversational Search",
    author = "Yang, Seunghan  and
      Lee, Juntae  and
      Bang, Jihwan  and
      Shim, Kyuhong  and
      Kim, Minsoo  and
      Chang, Simyung",
    editor = "Christodoulopoulos, Christos  and
      Chakraborty, Tanmoy  and
      Rose, Carolyn  and
      Peng, Violet",
    booktitle = "Proceedings of the 2025 Conference on Empirical Methods in Natural Language Processing",
    month = nov,
    year = "2025",
    address = "Suzhou, China",
    publisher = "Association for Computational Linguistics",
    url = "https://aclanthology.org/2025.emnlp-main.602/",
    doi = "10.18653/v1/2025.emnlp-main.602",
    pages = "11991--12003",
    ISBN = "979-8-89176-332-6"
}

@misc{bge_m3,
  title={BGE M3-Embedding: Multi-Lingual, Multi-Functionality, Multi-Granularity Text Embeddings Through Self-Knowledge Distillation},
  author={Chen, Jianlv and Xiao, Shitao and Zhang, Peitian and Luo, Kun and Lian, Defu and Liu, Zheng},
  year={2023},
  eprint={2309.07597},
  archivePrefix={arXiv},
  primaryClass={cs.CL}
}

@inproceedings{hyde,
  author       = {Luyu Gao and
                  Xueguang Ma and
                  Jimmy Lin and
                  Jamie Callan},
  editor       = {Anna Rogers and
                  Jordan L. Boyd{-}Graber and
                  Naoaki Okazaki},
  title        = {Precise Zero-Shot Dense Retrieval without Relevance Labels},
  booktitle    = {Proceedings of the 61st Annual Meeting of the Association for Computational
                  Linguistics (Volume 1: Long Papers), {ACL} 2023, Toronto, Canada,
                  July 9-14, 2023},
  pages        = {1762--1777},
  publisher    = {Association for Computational Linguistics},
  year         = {2023},
  url          = {https://doi.org/10.18653/v1/2023.acl-long.99},
  doi          = {10.18653/V1/2023.ACL-LONG.99},
  bibsource    = {dblp computer science bibliography, https://dblp.org}
}

@inproceedings{ambigqa,
  author       = {Sewon Min and
                  Julian Michael and
                  Hannaneh Hajishirzi and
                  Luke Zettlemoyer},
  editor       = {Bonnie Webber and
                  Trevor Cohn and
                  Yulan He and
                  Yang Liu},
  title        = {AmbigQA: Answering Ambiguous Open-domain Questions},
  booktitle    = {Proceedings of the 2020 Conference on Empirical Methods in Natural
                  Language Processing, {EMNLP} 2020, Online, November 16-20, 2020},
  pages        = {5783--5797},
  publisher    = {Association for Computational Linguistics},
  year         = {2020},
  url          = {https://doi.org/10.18653/v1/2020.emnlp-main.466},
  doi          = {10.18653/V1/2020.EMNLP-MAIN.466},
  bibsource    = {dblp computer science bibliography, https://dblp.org}
}

@inproceedings{followir,
  author       = {Orion Weller and
                  Benjamin Chang and
                  Sean MacAvaney and
                  Kyle Lo and
                  Arman Cohan and
                  Benjamin Van Durme and
                  Dawn J. Lawrie and
                  Luca Soldaini},
  editor       = {Luis Chiruzzo and
                  Alan Ritter and
                  Lu Wang},
  title        = {FollowIR: Evaluating and Teaching Information Retrieval Models to
                  Follow Instructions},
  booktitle    = {Proceedings of the 2025 Conference of the Nations of the Americas
                  Chapter of the Association for Computational Linguistics: Human Language
                  Technologies, {NAACL} 2025 - Volume 1: Long Papers, Albuquerque, New
                  Mexico, USA, April 29 - May 4, 2025},
  pages        = {11926--11942},
  publisher    = {Association for Computational Linguistics},
  year         = {2025},
  url          = {https://doi.org/10.18653/v1/2025.naacl-long.597},
  doi          = {10.18653/V1/2025.NAACL-LONG.597},
  bibsource    = {dblp computer science bibliography, https://dblp.org}
}

@article{conv_search_survey,
  author       = {Fengran Mo and
                  Kelong Mao and
                  Ziliang Zhao and
                  Hongjin Qian and
                  Haonan Chen and
                  Yiruo Cheng and
                  Xiaoxi Li and
                  Yutao Zhu and
                  Zhicheng Dou and
                  Jian{-}Yun Nie},
  title        = {A Survey of Conversational Search},
  journal      = {{ACM} Trans. Inf. Syst.},
  volume       = {43},
  number       = {6},
  pages        = {167:1--167:50},
  year         = {2025},
  url          = {https://doi.org/10.1145/3759453},
  doi          = {10.1145/3759453},
  bibsource    = {dblp computer science bibliography, https://dblp.org}
}

@inproceedings{intent_demonstrations_zhao,
  author       = {Ziliang Zhao and
                  Changle Qu and
                  Zhicheng Dou and
                  Haonan Chen and
                  Jiajie Jin},
  editor       = {Luiza Antonie and
                  Jian Pei and
                  Xiaohui Yu and
                  Flavio Chierichetti and
                  Hady W. Lauw and
                  Yizhou Sun and
                  Srinivasan Parthasarathy},
  title        = {Retrieving Intent-covering Demonstrations for Clarification Generation
                  in Conversational Search Systems},
  booktitle    = {Proceedings of the 31st {ACM} {SIGKDD} Conference on Knowledge Discovery
                  and Data Mining, V.2, {KDD} 2025, Toronto ON, Canada, August 3-7,
                  2025},
  pages        = {3992--4001},
  publisher    = {{ACM}},
  year         = {2025},
  url          = {https://doi.org/10.1145/3711896.3737104},
  doi          = {10.1145/3711896.3737104},
  bibsource    = {dblp computer science bibliography, https://dblp.org}
}

@article{reasonir,
  author       = {Rulin Shao and
                  Rui Qiao and
                  Varsha Kishore and
                  Niklas Muennighoff and
                  Xi Victoria Lin and
                  Daniela Rus and
                  Bryan Kian Hsiang Low and
                  Sewon Min and
                  Wen{-}tau Yih and
                  Pang Wei Koh and
                  Luke Zettlemoyer},
  title        = {ReasonIR: Training Retrievers for Reasoning Tasks},
  journal      = {CoRR},
  volume       = {abs/2504.20595},
  year         = {2025},
  url          = {https://doi.org/10.48550/arXiv.2504.20595},
  doi          = {10.48550/ARXIV.2504.20595},
  eprinttype    = {arXiv},
  eprint       = {2504.20595},
  bibsource    = {dblp computer science bibliography, https://dblp.org}
}

@inproceedings{brightbench,
  author       = {Hongjin Su and
                  Howard Yen and
                  Mengzhou Xia and
                  Weijia Shi and
                  Niklas Muennighoff and
                  Han{-}yu Wang and
                  Haisu Liu and
                  Quan Shi and
                  Zachary S. Siegel and
                  Michael Tang and
                  Ruoxi Sun and
                  Jinsung Yoon and
                  Sercan {\"{O}}. Arik and
                  Danqi Chen and
                  Tao Yu},
  title        = {{BRIGHT:} {A} Realistic and Challenging Benchmark for Reasoning-Intensive
                  Retrieval},
  booktitle    = {The Thirteenth International Conference on Learning Representations,
                  {ICLR} 2025, Singapore, April 24-28, 2025},
  publisher    = {OpenReview.net},
  year         = {2025},
  url          = {https://openreview.net/forum?id=ykuc5q381b},
  bibsource    = {dblp computer science bibliography, https://dblp.org}
}

@article{infonce,
  author       = {Gautier Izacard and
                  Mathilde Caron and
                  Lucas Hosseini and
                  Sebastian Riedel and
                  Piotr Bojanowski and
                  Armand Joulin and
                  Edouard Grave},
  title        = {Unsupervised Dense Information Retrieval with Contrastive Learning},
  journal      = {Trans. Mach. Learn. Res.},
  volume       = {2022},
  year         = {2022},
  url          = {https://openreview.net/forum?id=jKN1pXi7b0},
  bibsource    = {dblp computer science bibliography, https://dblp.org}
}

@article{gircse,
  author       = {Yu{-}Che Tsai and
                  Kuan{-}Yu Chen and
                  Yuan{-}Chi Li and
                  Yuan{-}Hao Chen and
                  Ching{-}Yu Tsai and
                  Shou{-}De Lin},
  title        = {Let LLMs Speak Embedding Languages: Generative Text Embeddings via
                  Iterative Contrastive Refinement},
  journal      = {CoRR},
  volume       = {abs/2509.24291},
  year         = {2025},
  url          = {https://doi.org/10.48550/arXiv.2509.24291},
  doi          = {10.48550/ARXIV.2509.24291},
  eprinttype    = {arXiv},
  eprint       = {2509.24291},
  bibsource    = {dblp computer science bibliography, https://dblp.org}
}

@article{qwen3embedding,
  author       = {Yanzhao Zhang and
                  Mingxin Li and
                  Dingkun Long and
                  Xin Zhang and
                  Huan Lin and
                  Baosong Yang and
                  Pengjun Xie and
                  An Yang and
                  Dayiheng Liu and
                  Junyang Lin and
                  Fei Huang and
                  Jingren Zhou},
  title        = {Qwen3 Embedding: Advancing Text Embedding and Reranking Through Foundation
                  Models},
  journal      = {CoRR},
  volume       = {abs/2506.05176},
  year         = {2025},
  url          = {https://doi.org/10.48550/arXiv.2506.05176},
  doi          = {10.48550/ARXIV.2506.05176},
  eprinttype    = {arXiv},
  eprint       = {2506.05176},
  bibsource    = {dblp computer science bibliography, https://dblp.org}
}

@article{gte,
  author       = {Zehan Li and
                  Xin Zhang and
                  Yanzhao Zhang and
                  Dingkun Long and
                  Pengjun Xie and
                  Meishan Zhang},
  title        = {Towards General Text Embeddings with Multi-stage Contrastive Learning},
  journal      = {CoRR},
  volume       = {abs/2308.03281},
  year         = {2023},
  url          = {https://doi.org/10.48550/arXiv.2308.03281},
  doi          = {10.48550/ARXIV.2308.03281},
  eprinttype    = {arXiv},
  eprint       = {2308.03281},
  bibsource    = {dblp computer science bibliography, https://dblp.org}
}

@inproceedings{nvembed,
  author       = {Chankyu Lee and
                  Rajarshi Roy and
                  Mengyao Xu and
                  Jonathan Raiman and
                  Mohammad Shoeybi and
                  Bryan Catanzaro and
                  Wei Ping},
  title        = {NV-Embed: Improved Techniques for Training LLMs as Generalist Embedding
                  Models},
  booktitle    = {The Thirteenth International Conference on Learning Representations,
                  {ICLR} 2025, Singapore, April 24-28, 2025},
  publisher    = {OpenReview.net},
  year         = {2025},
  url          = {https://openreview.net/forum?id=lgsyLSsDRe},
  bibsource    = {dblp computer science bibliography, https://dblp.org}
}

@inproceedings{e5mistral,
  author       = {Liang Wang and
                  Nan Yang and
                  Xiaolong Huang and
                  Linjun Yang and
                  Rangan Majumder and
                  Furu Wei},
  editor       = {Lun{-}Wei Ku and
                  Andre Martins and
                  Vivek Srikumar},
  title        = {Improving Text Embeddings with Large Language Models},
  booktitle    = {Proceedings of the 62nd Annual Meeting of the Association for Computational
                  Linguistics (Volume 1: Long Papers), {ACL} 2024, Bangkok, Thailand,
                  August 11-16, 2024},
  pages        = {11897--11916},
  publisher    = {Association for Computational Linguistics},
  year         = {2024},
  url          = {https://doi.org/10.18653/v1/2024.acl-long.642},
  doi          = {10.18653/V1/2024.ACL-LONG.642},
  bibsource    = {dblp computer science bibliography, https://dblp.org}
}

@article{gpt_4o_system_card,
  title={Gpt-4o system card},
  author={Hurst, Aaron and Lerer, Adam and Goucher, Adam P and Perelman, Adam and Ramesh, Aditya and Clark, Aidan and Ostrow, AJ and Welihinda, Akila and Hayes, Alan and Radford, Alec and others},
  journal={arXiv preprint arXiv:2410.21276},
  year={2024}
}

@inproceedings{bert,
  title={BERT: Pre-training of Deep Bidirectional Transformers for Language Understanding},
  author={Kenton, Jacob Devlin Ming-Wei Chang and Toutanova, Lee Kristina},
  booktitle={NAACL-HLT},
  pages={4171--4186},
  year={2019}
}

@inproceedings{dpr,
  title={Dense Passage Retrieval for Open-Domain Question Answering},
  author={Karpukhin, Vladimir and Oguz, Barlas and Min, Sewon and Lewis, Patrick and Wu, Ledell and Edunov, Sergey and Chen, Danqi and Yih, Wen-tau},
  booktitle={EMNLP},
  pages={6769--6781},
  year={2020}
}

@inproceedings{rag,
  author       = {Patrick S. H. Lewis and
                  Ethan Perez and
                  Aleksandra Piktus and
                  Fabio Petroni and
                  Vladimir Karpukhin and
                  Naman Goyal and
                  Heinrich K{\"{u}}ttler and
                  Mike Lewis and
                  Wen-tau Yih and
                  Tim Rockt{\"{a}}schel and
                  Sebastian Riedel and
                  Douwe Kiela},
  
  title        = {Retrieval-Augmented Generation for Knowledge-Intensive {NLP} Tasks},
  booktitle    = {Advances in Neural Information Processing Systems 33: Annual Conference
                  on Neural Information Processing Systems 2020, NeurIPS 2020, December
                  6-12, 2020, virtual},
  year         = {2020},
  url          = {https://proceedings.neurips.cc/paper/2020/hash/6b493230205f780e1bc26945df7481e5-Abstract.html},
  bibsource    = {dblp computer science bibliography, https://dblp.org}
}

@inproceedings{hotpotqa,
    title = "{H}otpot{QA}: A Dataset for Diverse, Explainable Multi-hop Question Answering",
    author = "Yang, Zhilin  and
      Qi, Peng  and
      Zhang, Saizheng  and
      Bengio, Yoshua  and
      Cohen, William  and
      Salakhutdinov, Ruslan  and
      Manning, Christopher D.",
    booktitle = "EMNLP",
    month = oct # "-" # nov,
    year = "2018",
    address = "Brussels, Belgium",
    publisher = "Association for Computational Linguistics",
    url = "https://aclanthology.org/D18-1259",
    doi = "10.18653/v1/D18-1259",
    pages = "2369--2380",
}

@misc{ragsurvey,
      title={Retrieval-Augmented Generation for Large Language Models: A Survey}, 
      author={Yunfan Gao and Yun Xiong and Xinyu Gao and Kangxiang Jia and Jinliu Pan and Yuxi Bi and Yi Dai and Jiawei Sun and Qianyu Guo and Meng Wang and Haofen Wang},
      year={2024},
      eprint={2312.10997},
      archivePrefix={arXiv},
      primaryClass={cs.CL}
}

@article{llm4ir,
  author       = {Yutao Zhu and
                  Huaying Yuan and
                  Shuting Wang and
                  Jiongnan Liu and
                  Wenhan Liu and
                  Chenlong Deng and
                  Zhicheng Dou and
                  Ji{-}Rong Wen},
  title        = {Large Language Models for Information Retrieval: {A} Survey},
  journal      = {CoRR},
  volume       = {abs/2308.07107},
  year         = {2023},
  url          = {https://doi.org/10.48550/arXiv.2308.07107},
  doi          = {10.48550/ARXIV.2308.07107},
  eprinttype    = {arXiv},
  eprint       = {2308.07107},
  bibsource    = {dblp computer science bibliography, https://dblp.org}
}

@article{e5,
  author       = {Liang Wang and
                  Nan Yang and
                  Xiaolong Huang and
                  Binxing Jiao and
                  Linjun Yang and
                  Daxin Jiang and
                  Rangan Majumder and
                  Furu Wei},
  title        = {Text Embeddings by Weakly-Supervised Contrastive Pre-training},
  journal      = {CoRR},
  volume       = {abs/2212.03533},
  year         = {2022},
  url          = {https://doi.org/10.48550/arXiv.2212.03533},
  doi          = {10.48550/ARXIV.2212.03533},
  eprinttype    = {arXiv},
  eprint       = {2212.03533},
  bibsource    = {dblp computer science bibliography, https://dblp.org}
}

@article{llm-survey,
  author       = {Wayne Xin Zhao and
                  Kun Zhou and
                  Junyi Li and
                  Tianyi Tang and
                  Xiaolei Wang and
                  Yupeng Hou and
                  Yingqian Min and
                  Beichen Zhang and
                  Junjie Zhang and
                  Zican Dong and
                  Yifan Du and
                  Chen Yang and
                  Yushuo Chen and
                  Zhipeng Chen and
                  Jinhao Jiang and
                  Ruiyang Ren and
                  Yifan Li and
                  Xinyu Tang and
                  Zikang Liu and
                  Peiyu Liu and
                  Jian-Yun Nie and
                  Ji-Rong Wen},
  title        = {A Survey of Large Language Models},
  journal      = {CoRR},
  volume       = {abs/2303.18223},
  year         = {2023},
  url          = {https://doi.org/10.48550/arXiv.2303.18223},
  doi          = {10.48550/ARXIV.2303.18223},
  eprinttype    = {arXiv},
  eprint       = {2303.18223},
  bibsource    = {dblp computer science bibliography, https://dblp.org}
}

@article{ir_meets_llm,
  author       = {Qingyao Ai and
                  Ting Bai and
                  Zhao Cao and
                  Yi Chang and
                  Jiawei Chen and
                  Zhumin Chen and
                  Zhiyong Cheng and
                  Shoubin Dong and
                  Zhicheng Dou and
                  Fuli Feng and
                  Shen Gao and
                  Jiafeng Guo and
                  Xiangnan He and
                  Yanyan Lan and
                  Chenliang Li and
                  Yiqun Liu and
                  Ziyu Lyu and
                  Weizhi Ma and
                  Jun Ma and
                  Zhaochun Ren and
                  Pengjie Ren and
                  Zhiqiang Wang and
                  Mingwen Wang and
                  Ji-Rong Wen and
                  Le Wu and
                  Xin Xin and
                  Jun Xu and
                  Dawei Yin and
                  Peng Zhang and
                  Fan Zhang and
                  Weinan Zhang and
                  Min Zhang and
                  Xiaofei Zhu},
  title        = {Information Retrieval meets Large Language Models: {A} strategic report
                  from Chinese {IR} community},
  journal      = {{AI} Open},
  volume       = {4},
  pages        = {80--90},
  year         = {2023},
  url          = {https://doi.org/10.1016/j.aiopen.2023.08.001},
  doi          = {10.1016/J.AIOPEN.2023.08.001},
  bibsource    = {dblp computer science bibliography, https://dblp.org}
}

@article{COT,
  author       = {Jason Wei and
                  Xuezhi Wang and
                  Dale Schuurmans and
                  Maarten Bosma and
                  Ed H. Chi and
                  Quoc Le and
                  Denny Zhou},
  title        = {Chain of Thought Prompting Elicits Reasoning in Large Language Models},
  journal      = {CoRR},
  volume       = {abs/2201.11903},
  year         = {2022},
  url          = {https://arxiv.org/abs/2201.11903},
  eprinttype    = {arXiv},
  eprint       = {2201.11903},
  bibsource    = {dblp computer science bibliography, https://dblp.org}
}

@article{2402_gritlm,
  author       = {Niklas Muennighoff and
                  Hongjin Su and
                  Liang Wang and
                  Nan Yang and
                  Furu Wei and
                  Tao Yu and
                  Amanpreet Singh and
                  Douwe Kiela},
  title        = {Generative Representational Instruction Tuning},
  journal      = {CoRR},
  volume       = {abs/2402.09906},
  year         = {2024},
  url          = {https://doi.org/10.48550/arXiv.2402.09906},
  doi          = {10.48550/ARXIV.2402.09906},
  eprinttype    = {arXiv},
  eprint       = {2402.09906},
  bibsource    = {dblp computer science bibliography, https://dblp.org}
}

@inproceedings{vllm,
  author       = {Woosuk Kwon and
                  Zhuohan Li and
                  Siyuan Zhuang and
                  Ying Sheng and
                  Lianmin Zheng and
                  Cody Hao Yu and
                  Joseph Gonzalez and
                  Hao Zhang and
                  Ion Stoica},
  editor       = {Jason Flinn and
                  Margo I. Seltzer and
                  Peter Druschel and
                  Antoine Kaufmann and
                  Jonathan Mace},
  title        = {Efficient Memory Management for Large Language Model Serving with
                  PagedAttention},
  booktitle    = {Proceedings of the 29th Symposium on Operating Systems Principles,
                  {SOSP} 2023, Koblenz, Germany, October 23-26, 2023},
  pages        = {611--626},
  publisher    = {{ACM}},
  year         = {2023},
  url          = {https://doi.org/10.1145/3600006.3613165},
  doi          = {10.1145/3600006.3613165},
  bibsource    = {dblp computer science bibliography, https://dblp.org}
}

@inproceedings{2wiki,
  author       = {Xanh Ho and
                  Anh{-}Khoa Duong Nguyen and
                  Saku Sugawara and
                  Akiko Aizawa},
  editor       = {Donia Scott and
                  N{\'{u}}ria Bel and
                  Chengqing Zong},
  title        = {Constructing {A} Multi-hop {QA} Dataset for Comprehensive Evaluation
                  of Reasoning Steps},
  booktitle    = {Proceedings of the 28th International Conference on Computational
                  Linguistics, {COLING} 2020, Barcelona, Spain (Online), December 8-13,
                  2020},
  pages        = {6609--6625},
  publisher    = {International Committee on Computational Linguistics},
  year         = {2020},
  url          = {https://doi.org/10.18653/v1/2020.coling-main.580},
  doi          = {10.18653/V1/2020.COLING-MAIN.580},
  bibsource    = {dblp computer science bibliography, https://dblp.org}
}

@inproceedings{bge,
  author       = {Shitao Xiao and
                  Zheng Liu and
                  Peitian Zhang and
                  Niklas Muennighoff and
                  Defu Lian and
                  Jian{-}Yun Nie},
  editor       = {Grace Hui Yang and
                  Hongning Wang and
                  Sam Han and
                  Claudia Hauff and
                  Guido Zuccon and
                  Yi Zhang},
  title        = {C-Pack: Packed Resources For General Chinese Embeddings},
  booktitle    = {Proceedings of the 47th International {ACM} {SIGIR} Conference on
                  Research and Development in Information Retrieval, {SIGIR} 2024, Washington
                  DC, USA, July 14-18, 2024},
  pages        = {641--649},
  publisher    = {{ACM}},
  year         = {2024},
  url          = {https://doi.org/10.1145/3626772.3657878},
  doi          = {10.1145/3626772.3657878},
  bibsource    = {dblp computer science bibliography, https://dblp.org}
}
\clearpage

\end{document}